\documentclass[lettersize,journal]{IEEEtran}
\usepackage{algorithmic}
\usepackage{algorithm}
\usepackage{array}
\usepackage{textcomp}
\usepackage{stfloats}
\usepackage{url}
\usepackage{verbatim}
\usepackage{graphicx}
\usepackage{cite}
\usepackage[hidelinks]{hyperref}
\usepackage{subcaption}
\usepackage[fleqn]{amsmath}
\usepackage{booktabs}
\usepackage{multirow}
\usepackage{tikz}
\usetikzlibrary{calc}
\usepackage{xcolor}
\DeclareMathOperator*{\argmin}{argmin}

\newcommand{\etal}{~\textit{et~al. }}
\newcommand{\mysection}[1]{\leavevmode\newline\noindent\textbf{#1}\hspace{2ex}}
\newcommand{\myunderbracket}[3][0mm]{%
    \begin{tikzpicture}[anchor=base, baseline]
        \node (mytarget) [inner sep=0pt] {#2};
        \draw [line width=1pt]
            let \p1 = (mytarget.south west),
                \p2 = (mytarget.south east),
                in
                (\x1 + 1pt, \y1) --
                (\x1 + 1pt, \y1 - 2mm - #1) --
                (\x2 - 1pt, \y2 - 2mm - #1) --
                (\x2 - 1pt, \y2);
        \node [anchor=center, fill=white, rounded corners] at ([yshift=-2mm - #1] mytarget.south) {#3};
    \end{tikzpicture}%
}

\begin{document}
\bstctlcite{bstctl:nodash} 

\title{Polarimetric Light Transport Analysis \\for Specular Inter-reflection}


\author{Ryota Maeda and Shinsaku Hiura
\thanks{R. Maeda and S. Hiura are affiliated with Graduate School of Engineering, University of Hyogo (e-mail: maeda.ryota.elerac@gmail.com; hiura@eng.u-hyogo.ac.jp).}}



\maketitle

\begin{abstract}
Polarization is well known for its ability to decompose diffuse and specular reflections. However, the existing decomposition methods only focus on direct reflection and overlook multiple reflections, especially specular inter-reflection. In this paper, we propose a novel decomposition method for handling specular inter-reflection of metal objects by using a unique polarimetric feature: the rotation direction of linear polarization. This rotation direction serves as a discriminative factor between direct and inter-reflection on specular surfaces. To decompose the reflectance components, we actively rotate the linear polarization of incident light and analyze the rotation direction of the reflected light. We evaluate our method using both synthetic and real data, demonstrating its effectiveness in decomposing specular inter-reflections of metal objects. Furthermore, we demonstrate that our method can be combined with other decomposition methods for a detailed analysis of light transport. As a practical application, we show its effectiveness in improving the accuracy of 3D measurement against strong specular inter-reflection.
\end{abstract}


\section{Introduction}

\IEEEPARstart{T}{he} 
analysis of light transport is an important task in computer vision, allowing us to understand the real-world scene through various optical phenomena, such as diffuse and specular reflection, subsurface scattering, and inter-reflection. These phenomena repeatedly occur in intricate geometry and materials, resulting in a complex mixture of light observations. Sometimes, these mixed lights lead to difficulties in subsequent tasks, such as 3D measurement and material classification. Therefore, analyzing the mixed light transport is essential for understanding the scene.

There are numerous different techniques to decompose light transport components. These techniques use the relationship between reflection components and optical properties, such as the dichromatic reflectance model~\cite{shafer1985using, klinker1988measurement, sato1994temporal, sato1997object, tan2008separating, kim2013specular}, spatial frequency response~\cite{nayar2006fast}, geometric constraint~\cite{o2012primal, o20143d, o2015homogeneous}, far infrared light~\cite{tanaka2019time} and polarization~\cite{wolff1993constraining, muller1996elimination, debevec2000acquiring, schechner2001instant, schechner2005recovery, ma2007rapid, ghosh2010circularly}. These different physical cues target different optical phenomena, and they complement each other. Our polarization-based method falls within that category. 

Polarization is a renowned physical cue to decompose diffuse and specular reflection components. Numerous polarization-based methods exploit the behavior of reflection, wherein a diffuse surface makes the incident light unpolarized while a specular surface preserves its polarization~\cite{wolff1993constraining, muller1996elimination, debevec2000acquiring, schechner2001instant, schechner2005recovery, ma2007rapid, ghosh2010circularly}. 
These methods assume that all indirect reflection components are unpolarized, similar to diffuse surface reflection or subsurface scattering.
However, this assumption is not always~true because inter-reflection on specular surfaces preserves the polarization of the incident light. Therefore, the existing methods are not capable of explicitly handling specular inter-reflection.

\begin{figure}[tb]
    \centering
    \includegraphics[width=\hsize]{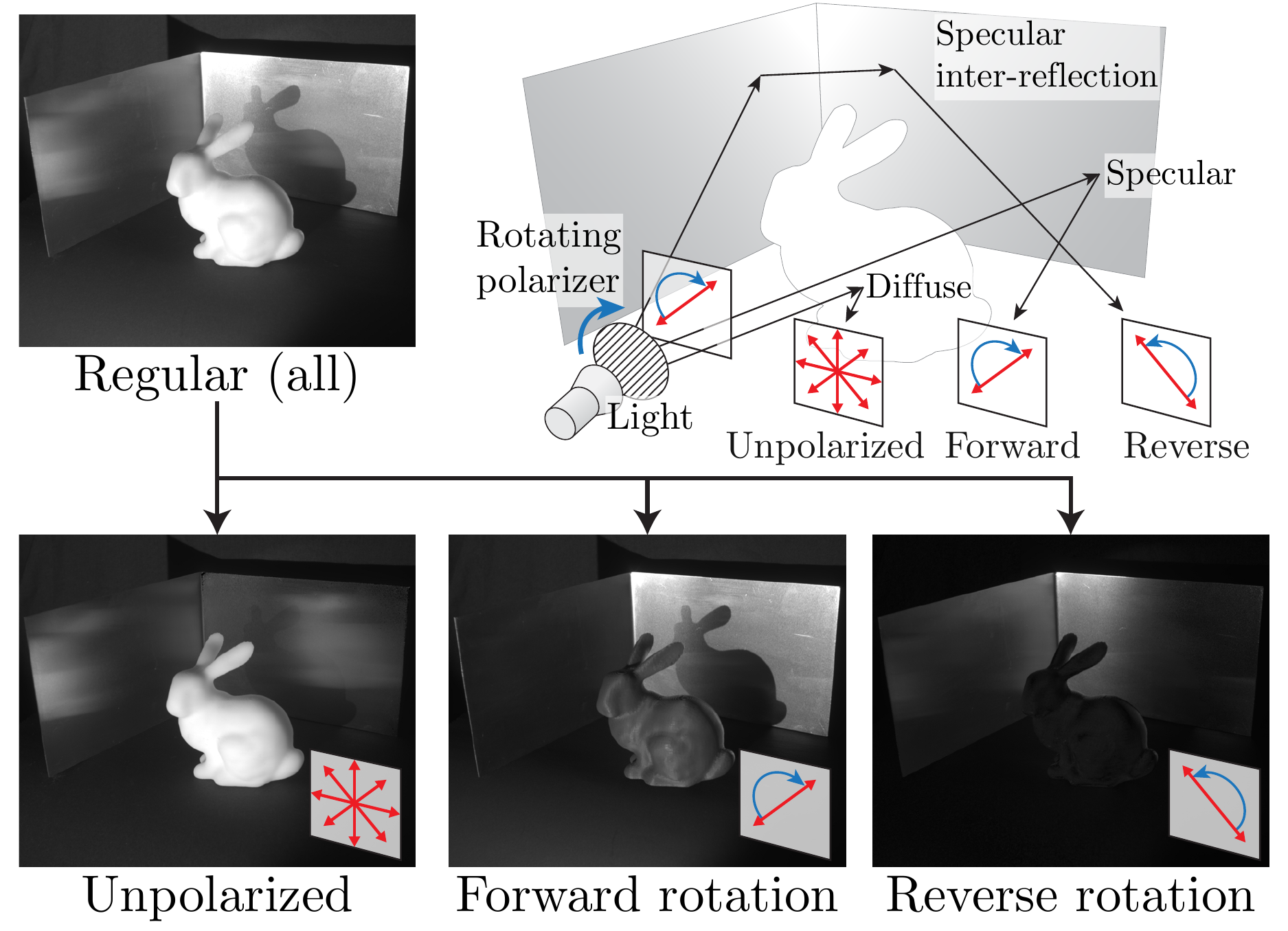}
    \caption{\textbf{Overview of reflectance components decomposition based on the rotation of linear polarization.} Our method can decompose polarization states of the reflected light into three components: \textit{Unpolarized}, \textit{Forward rotation}, and \textit{Reverse rotation}.}
    \label{fig:teaser}
\end{figure}

In this paper, we propose a novel polarization-based light transport decomposition method, especially for specular inter-reflection. Our method is based on the observation of specular inter-reflection that, when the polarization plane of incident light rotates, the polarization plane of the reflected light rotates in the opposite direction. We exploit this observation as a useful cue to decompose mixed light. To capture the rotation of polarization, we use a pair of linear polarizers in front of the light source and camera, and the camera captures the polarization state of the reflected light by rotating both polarizers. As shown in Fig.~\ref{fig:teaser}, we decompose the reflected light into three components: \textit{Unpolarized}, \textit{Forward rotation}, \textit{Reverse rotation}. These components correspond to the diffuse reflection, specular reflection, and specular inter-reflection, respectively. 
As our method utilizes a polarization cue, we can easily combine it with existing methods that use a different physical cue. This combination allows us to decompose high- and low-frequency specular inter-reflection that cannot be revealed by existing methods. Furthermore, with our method, we can measure the 3D shape in the presence of specular inter-reflection. 

To summarize, our main contributions are:
\begin{itemize}
    \item We present a novel polarization-based light transport decomposition method, especially for specular inter-reflection (Fig.~\ref{fig:result_decomposition}), which is challenging in previous methods (Fig.~\ref{fig:result_vs_nayar}).
    \item We show that our method can be combined with existing methods that use a different physical cue because our method only relies on polarization (Fig.~\ref{fig:result_with_nayar}). 
    \item We show a practical application of our method: 3D measurement with a projector-camera system for metal objects (Fig.~\ref{fig:result_3d_measurement}). With our method, we can measure the 3D shape in the presence of specular inter-reflection. 
\end{itemize}
Similar to other existing methods, our method also has several limitations, which are discussed in Section \ref{sec:limitation}.

\begin{figure*}[tbh]
    \centering
    \begin{minipage}[c]{\hsize}
        \centering
        \begin{minipage}[c]{0.333\hsize}
            \centering
            \includegraphics[height=0.96\hsize]{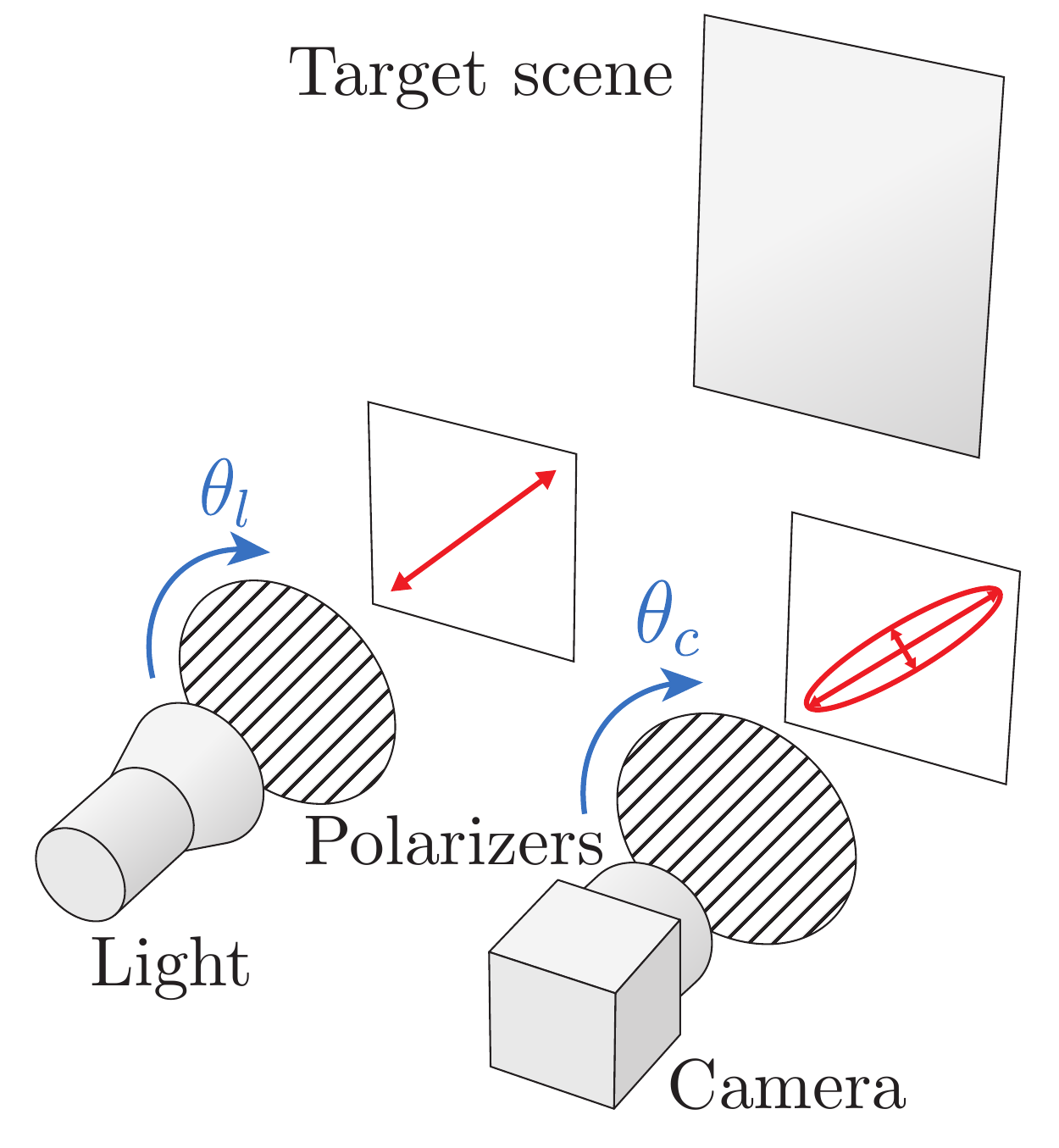}
            \subcaption{Capturing configuration}
        \end{minipage}%
        \begin{minipage}[c]{0.333\hsize}
            \centering
            \includegraphics[height=0.96\hsize]{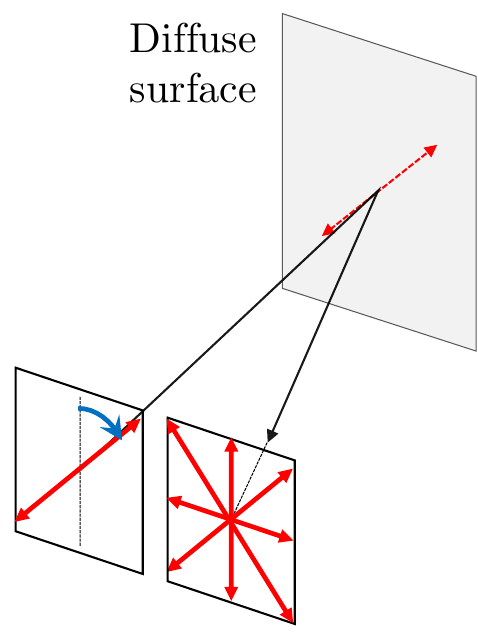}
            \subcaption{Unpolarized}
        \end{minipage}
    \end{minipage}
    \begin{minipage}[c]{\hsize}
        \centering
        \begin{minipage}[c]{0.333\hsize}
            \centering
            \includegraphics[height=0.96\hsize]{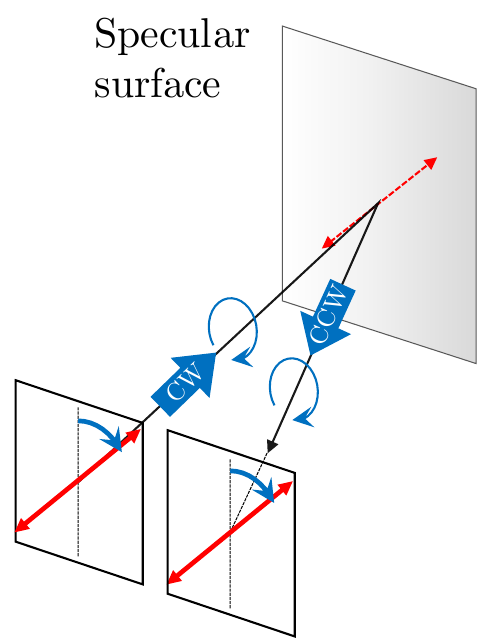}
            \\ \vphantom{{\small HorizontalOblique}} \\
            \subcaption{Forward rotation}
        \end{minipage}%
        \begin{minipage}[c]{0.666\hsize}
            \centering
            \begin{minipage}[c]{0.5\hsize}
                \centering
                \includegraphics[height=0.96\hsize]{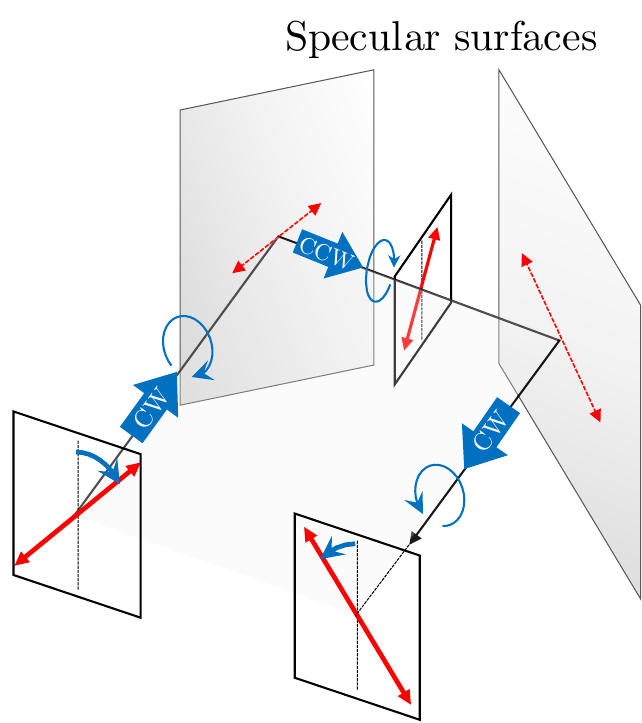}
                {\small Horizontal}
            \end{minipage}%
            \begin{minipage}[c]{0.5\hsize}
                \centering
                \includegraphics[height=0.94\hsize]{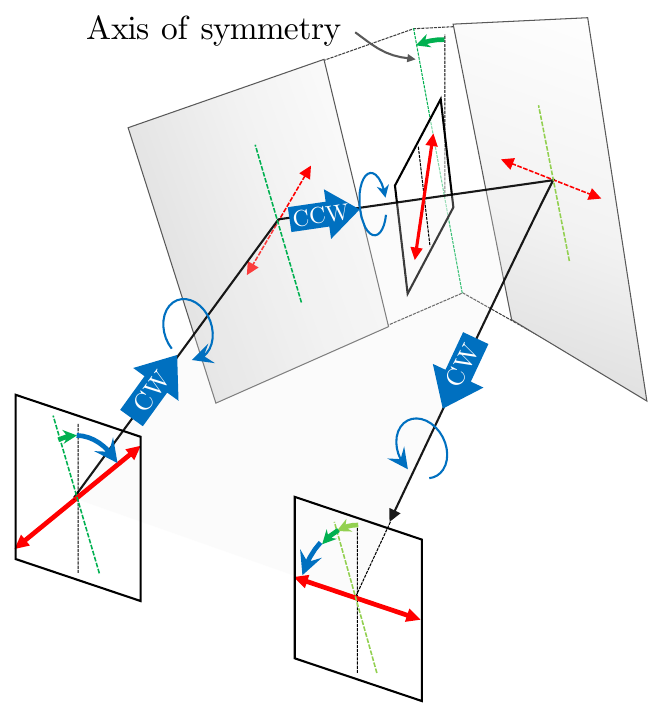}
                {\small Oblique}
            \end{minipage}
            \subcaption{Reverse rotation}
        \end{minipage}
    \end{minipage}

    \caption{\textbf{Polarization states of reflected light relative to incident light's polarization plane rotation.} (a) We illuminate the target scene with a light source with a polarizer and capture the reflected light with a camera with a polarizer. (b) In diffuse reflection or subsurface scattering, the reflected light is mostly unpolarized. (c) In direct specular reflection, the polarization plane of the reflected light rotates in the same direction as the incident light. (d) In specular inter-reflection, on the other hand, the polarization plane of the reflected light rotates in the opposite direction from the incident light. The phase of the polarization plane varies depending on the orientation of the reflection surfaces (compare horizontal and oblique).}
    \label{fig:pol_rot_states}
\end{figure*}

\section{Related Work}
\subsection{Reflection and Light Transport Analysis}
The decomposition of reflection components is an essential task for computer vision and graphics, so many researchers have explored various methods. In essence, these methods exploit the relationship between reflection components and physical optical phenomena. 
Shafer\etal~\cite{shafer1985using} separate diffuse and specular reflection based on the dichromatic reflectance model. This model incorporates the color difference due to diffuse and specular reflection. 
Several reflection decomposition methods utilize the spatially high-frequency light pattern. Nayar\etal~\cite{nayar2006fast} proposed a high-frequency spatial patterns method to separate direct and global components. This principle is applied to robust 3D scanning methods against inter-reflection and subsurface scattering~\cite{couture2011unstructured, gupta2012micro, gupta2013practical}. 
Some methods exploit the geometric constraints of the camera and light source. O'Toole\etal proposed primal-dual coding~\cite{o2012primal} using a coaxial projector-camera setup, and his follow-up work uses epipolar constraints~\cite{o20143d, o2015homogeneous}.
Other methods decompose reflection components by far infrared
light~\cite{tanaka2019time} or the time-of-flight approach~\cite{wu2014decomposing}.

Polarization is also an essential cue for the decomposition of reflection. It is well known that the degree of polarization differs in diffuse and specular reflection components. Several methods have been developed to analyze the polarization states of reflected light~\cite{wolff1993constraining, muller1996elimination, debevec2000acquiring, ma2007rapid, ghosh2010circularly}, and estimate the 3D shape of translucent objects, which causes subsurface scattering~\cite{chen2007polarization}.
There are still few studies on analyzing specular inter-reflection using polarized light. Wallance\etal distinguish direct and inter-reflection of metallic objects ~\cite{wallace1999improving}. However, they worked not to decompose but just distinguish them. In addition, their experimental results showed only handling horizontal inter-reflections and did not account for inter-reflections in arbitrary directions. In contrast, we decompose the reflection components, which can apply to various inter-reflection directions.

\subsection{Polarization in Computer Vision}
In computer vision, polarization has been exploited not only for the separation of diffuse and specular reflection but also for various imaging techniques such as 3D shape recovery for both opaque~\cite{miyazaki2003polarization, kadambi2015polarized, ding2021polarimetric} and transparent~\cite{miyazaki2004transparent} objects, dehazing~\cite{schechner2001instant, treibitz2008active}, and segmentation for transparent objects~\cite{kalra2020deep}. These studies were conducted under uncontrollable environment light or known light sources with fixed polarizers.

Recently, Baek\etal explored the polarimetric relationship between the incident and reflected light for various imaging applications~\cite{baek2018simultaneous, baek2020image, baek2021polarimetric, baek2022all}. Similar to his series of works, we measure the polarimetric information using rotating polarizers attached to both light source and camera sides. In particular, Baek and Heide briefly showed a measured result of the specular inter-reflection~\cite{baek2021polarimetric}, which implies the circular polarization change in specular inter-reflection. Our research delves deeper into this polarization change by explicitly defining the mathematical model and utilizing it for decomposing light transport with polarization cue alone.
\section{Principle}
\label{sec:principle}

\begin{figure*}[t]
    \centering
    \begin{minipage}{0.63582644\hsize}
        \centering
        \includegraphics[width=\hsize]{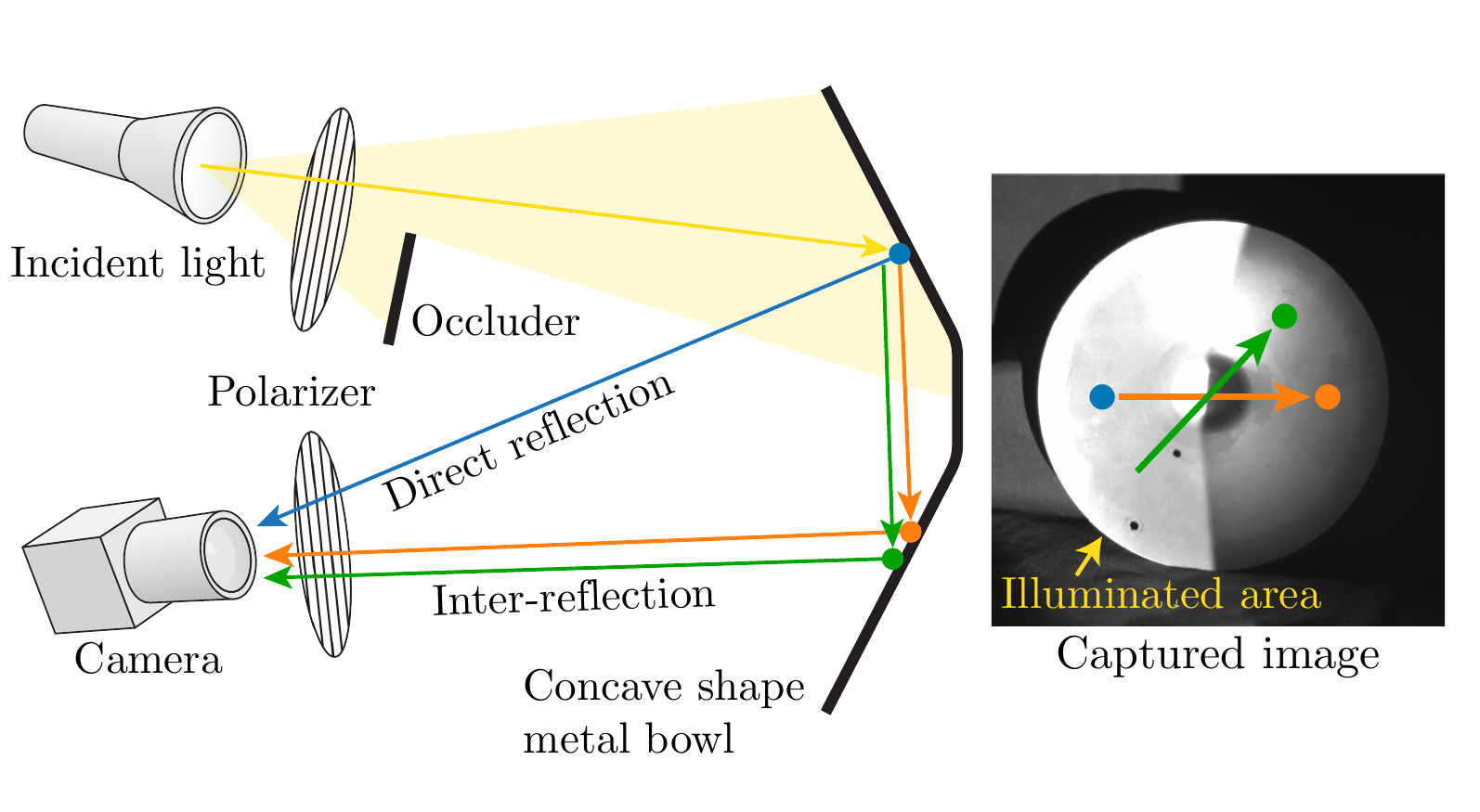}
        (a) Capturing setup with an occluder
    \end{minipage}%
    \begin{minipage}{0.36417356\hsize}
        \centering
        \includegraphics[width=\hsize]{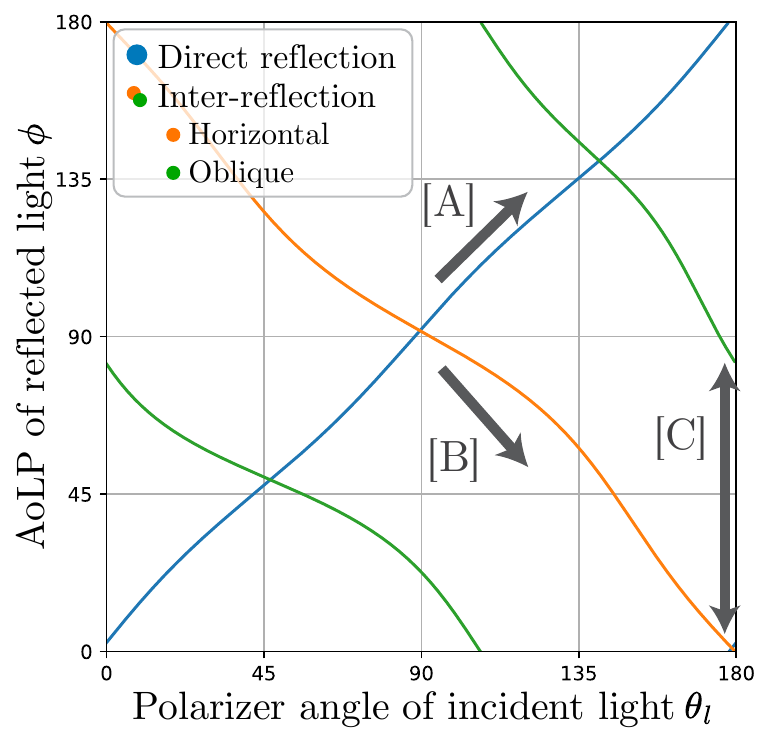}
        (b) The polarizer angle and measured AoLP
    \end{minipage}
    \caption{\textbf{Rotation of the angle of linear polarization in specular inter-reflection.} This measurement focuses on three different types of reflections: direct reflection, inter-reflection (horizontal), and inter-reflection (oblique). (a) The light source illuminates the concave-shaped metal bowl with the polarized light source. The occluder blocks the light, and the bowl is illuminated only on one side. This enables the camera to observe unmixed direct reflection and inter-reflection separately. Note that the occluder is specifically used to prevent mixing the light, and the later experiments are conducted without the occluder. (b) The plotted result of the AoLP of reflected light $\phi$ and the polarizer angle of light side $\theta_{l}$. We can see the different trends between direct and inter-reflection ([A] and [B]) and the phase difference between horizontal and oblique inter-reflection ([C]).}
    \label{fig:verify_rotation}
\end{figure*}

In this section, we introduce the essential polarization phenomenon in specular inter-reflection that forms the cornerstone of this study. Fig.~\ref{fig:pol_rot_states} illustrates our capturing configuration (a) and three polarization states that aim to decompose(b)(c)(d), and Fig.~\ref{fig:verify_rotation} shows the actual measurement of this phenomenon. As listed below, we explain polarization states in specular inter-reflection by referring to the intuitive interpretation in Fig.~\ref{fig:pol_rot_states} and the actual data in Fig.~\ref{fig:verify_rotation}. 
\begin{enumerate}
    \item \textit{Degree of linear polarization.} As shown in Fig.~\ref{fig:pol_rot_states}(b), the reflected light becomes unpolarized in diffuse reflection or subsurface scattering. In contrast, as illustrated in Fig.~\ref{fig:pol_rot_states}(c) and (d), polarization is preserved in specular reflection.
    \item \textit{Rotation direction.} As shown in Fig.~\ref{fig:pol_rot_states}(c)(d), the rotation direction of polarization of reflected light differs depending on the number of bounces. When the polarization is rotated in the clockwise (CW) direction, the reflection flips to counter-clockwise (CCW). Therefore, as shown in Fig.~\ref{fig:pol_rot_states}(c), in single bounce specular reflection, the polarization planes of the incident and reflected light rotates in the same direction from the camera viewpoint. On the other hand, in the case of the second bounce reflection, the rotation direction flips twice, and the reflected light becomes CW, resulting in a reversed rotation as shown in Fig.~\ref{fig:pol_rot_states}(d).
    We can see this opposite trend in real data, as indicated in [A] and [B] in Fig.~\ref{fig:verify_rotation}(b). With respect to the polarizer angle of the incident light, the angle of linear polarization (AoLP) of the direct reflection is linearly increasing, and the inter-reflection is decreasing\footnote{The plotted result is not perfectly linearly increasing or decreasing and is slightly distorted. This distortion is caused by various complex polarization phenomena, such as the fact that actual metal material yields elliptical polarization~\cite{collett2005field}.}. Note that, in the third reflection, the rotation direction flips three times and becomes CCW. It is the same as the first one.
    \item \textit{Phase difference.} As shown in Fig.~\ref{fig:pol_rot_states}(c)'s horizontal and oblique arrangements, the phase shift occurs in specular inter-reflection. We can confirm this phase difference in [C] of Fig.~\ref{fig:verify_rotation}(b). The amount of phase difference depends on the orientation of the light path. Note that we show a detailed visualization of the phase shift in Fig.\ref{fig:result_phase}.
\end{enumerate}

We can use the difference in rotation direction as a cue to distinguish three classes of reflection as shown in Fig.~\ref{fig:pol_rot_states}(b)(c)(d). However, in general, we observe not a single component but a mixture of them. Therefore, in the next section, we propose the decomposition method using a mathematical model of these three components.

\section{Method}
\label{sec:method}
In this section, we introduce our proposed decomposition method. In subsection A, we describe the basic relationship between observed intensity and polarization. Then, in subsection B, we formulate the mathematical model of three reflection components by incorporating the rotation direction of the plane of polarization. In subsection C, we show the decomposition method for three components. Finally, in subsection D, we discuss the minimum number of measurements required to apply the decomposition.

\subsection{Basic Observation of Polarization}
Before diving into our method, we briefly review the background of polarization. Although our proposed method uses a pair of linear polarizers in front of a light source and a camera side, we start with a single polarizer case. 
Suppose we measure the intensity of light with a camera that has a linear polarizer attached to the front. The camera measures the intensity $I(\theta_{c})$ at the angle of the linear polarizer $\theta_{c}$ which expressed as
\begin{equation}
    \label{equ:polarization-image-formation}
    I_{\rho, \phi}(\theta_{c}) = \frac{I_{\mathrm{amp}}}{2} \{\rho \cos{2(\theta_{c} - \phi) + 1} \},
\end{equation}
where $I_{\mathrm{amp}}$, $\rho$, and $\phi$ are the parameters that characterize the polarization of its light. $I_{\mathrm{amp}}$ is the intensity, $\rho$ is the degree of linear polarization (DoLP), and $\phi$ is the angle of linear polarization (AoLP). The DoLP represents how much the light is polarized with a value of $1$ indicating perfectly linear polarized light and $0$ indicating unpolarized light. The AoLP represents the angle of the plane of polarization light. The division by $2$ corresponds to the attenuation of the linear polarizer.

The equation above only represents the case of a single polarizer of a camera. In the next subsection, we will add a polarizer to the light source and extend the equation to consider the rotation direction of the polarization plane.

\subsection{Rotation Direction of Linear Polarization}
We extend the basic polarization expression Eq.~\ref{equ:polarization-image-formation} by incorporating the rotation direction of the plane of polarization. Our method defines the three types of polarization states illustrated in Fig.~\ref{fig:pol_rot_states}. As shown in Fig.~\ref{fig:pol_rot_states}(a), the scene is illuminated by a light source with a linear polarizer and captured by a camera with a linear polarizer. Both polarizers are to be rotated independently, and the camera observes the intensity $I(\theta_{c}, \theta_{l})$ at the angle of linear polarizers $\theta_{c}$ (camera side), $\theta_{l}$ (light source side). Then, we define the observed intensity as a mixture of three components, 
\begin{equation}
    \label{equ:sum_of_three_components}
    I(\theta_{c}, \theta_{l}) = I_{\textrm{unpolarized}}(\theta_{c}, \theta_{l}) + I_{\textrm{forward}}(\theta_{c}, \theta_{l}) + I_{\textrm{reverse}}(\theta_{c}, \theta_{l}),
\end{equation}
where $I_{\textrm{unpolarized}}$, $I_{\textrm{forward}}$, $I_{\textrm{reverse}}$ are the \textit{unpolarized}, \textit{forward rotation} and \textit{reverse rotation} components, respectively. We describe the detailed properties of these reflection components below.

\mysection{Unpolarized Components.}
As shown in Fig.~\ref{fig:pol_rot_states}(b), in the case of diffuse reflection and/or subsurface scattering, the reflected light becomes mostly unpolarized regardless of the polarization states of the incident light. The light enters the object and is reflected multiple times inside the medium so that the polarized light mixes in various directions and becomes unpolarized light (DoLP becomes almost zero). This unpolarized component is also partially present in rough specular reflection.
By using Eq.~\ref{equ:polarization-image-formation} with $\rho=0$, the \textit{unpolarized component} $I_{\textrm{unpolarized}}$ is observed as
\begin{equation}
    I_{\textrm{unpolarized}}(\theta_{c}, \theta_{l}) 
    = I_{\rho=0}(\theta_{c}) 
    = \frac{I_{U}}{2},
    \label{equ:unpolarized_component}
\end{equation}
where $I_{U}$ is the intensity of reflection. As the equation shows, $I_{\textrm{unpolarized}}$ is constant because the unpolarized component is not affected by the angle of the polarizers.

\mysection{Forward Rotation Components.}
The direct specular reflection mostly preserves the polarization of incident light, so it has a high DoLP value. When we rotate the polarization plane of the incident light (i.e., rotating the polarizer in front of the light source), the polarization plane of the reflected light is rotated in the forward direction as shown in Fig.~\ref{fig:pol_rot_states}(c). 
We define such reflected light as \textit{forward rotation components}, which corresponds to $\rho=1$ and $\phi=\theta_{l}+\phi_{F}$. 
Therefore, according to Eq.~\ref{equ:polarization-image-formation}, the intensity of forward rotation component $I_{\textrm{forward}}$ is observed as
\begin{equation}
    \begin{aligned}
        I_{\textrm{forward}}(\theta_{c}, \theta_{l}) 
        &= I_{\rho=1, \phi=\theta_{l}+\phi_{F}}(\theta_{c}) \\
        &= \frac{I_{F}}{2} \{\cos{2(\theta_{c} - \theta_{l} - \phi_{F})} + 1\},
    \end{aligned}
    \label{equ:forward_component}
\end{equation}
where $I_{F}$ represents the intensity of the forward rotation component, and $\phi_{F}$ is the phase typically close to zero.

\mysection{Reverse Rotation Components.}
We consider the second bounce specular inter-reflection. Even if it is reflected multiple times, the light still preserves the polarization (high DoLP value), similar to the single bounce case described before.
The difference between the direct- and the inter- reflection is the rotation direction of the plane of polarization. In specular inter-reflection, the polarization plane rotates reversely, as shown in Fig.~\ref{fig:pol_rot_states}(d). We define such reflected light as \textit{reverse rotation components}, which corresponds to $\rho=1$ and $\phi=-\theta_{l}+\phi_{R}$. Compared to the forward rotation components, the sign of $\theta_{l}$ is flipped. With Eq.~\ref{equ:polarization-image-formation}, the reverse rotation component $I_{\textrm{reverse}}$ is observed as 
\begin{equation}
    \begin{aligned}
        I_{\textrm{reverse}}(\theta_{c}, \theta_{l}) 
        &= I_{\rho=1, \phi=-\theta_{l}+\phi_{R}}(\theta_{c}) \\
        &= \frac{I_{R}}{2} \{\cos{2(\theta_{c} + \theta_{l} - \phi_{R})} + 1\},
    \end{aligned}
    \label{equ:reverse_component}
\end{equation}
where $I_{R}$ is the intensity of the reverse rotation component, and $\phi_{R}$ is the phase that depends on the arrangement of the equipment and objects, shown as [C] in Fig.~\ref{fig:verify_rotation}(b).

\subsection{Decomposition}
Based on the multiple observations of the intensities under different polarizer angles, we can decompose the mixed light into three components. We capture the static scene by rotating the linear polarizer of both the light source and camera sides and acquiring sequential images. For the $k$-th ($k=1, 2, ... , N$) image, let $I_{k}$ be the intensity, where the angle of the polarizer is $\theta_{c,k}$ (camera side) and $\theta_{l,k}$ (light source side). Then, we can estimate the parameters of each reflection component by solving the following equation.
\begin{equation}
    \hat{I_{U}}, \hat{I_{F}}, \hat{\phi_{F}}, \hat{I_{R}}, \hat{\phi_{R}} = \argmin_{I_{U},I_{F},\phi_{F},I_{R},\phi_{R}} \sum_{k=1}^{N}\left\{I_{k}- I(\theta_{c,k}, \theta_{l,k})\right\}^{2}.
    \label{equ:non-linear_optimization_problem}
\end{equation}

This equation can be solved by a linear least-square method, and we can determine the parameters in a closed form. We describe the details in the appendix.

\subsection{Minimum Number of Measurement}
\label{subsec:number_of_measurement}
The proposed decomposition method requires capturing multiple images with sufficient combinations of polarizer angles to find the five unknown parameters. We summarize the minimum number of images required for two cases of camera equipment. We show the proof in the Appendix.

\subsubsection{Conventional camera with polarizer} 
When using a polarizer with a conventional camera, both the polarizers on the light source and the camera must be rotated. In this case, we need to capture five images with non-degenerated combinations of polarizer angles, such as $\{(\theta_{c}, \theta_{l})\} = \{(0, 0), (45,45), (0,45), (45,0), (90,0)\}$.

\subsubsection{Polarization camera} 
Instead of using a conventional camera with a polarizer, we can use a polarization camera. This camera can acquire full linear polarization states in one shot, reducing the number of shots. In this case, we need to capture two images with 0 and 45 degrees of polarizer angle on the light source.

\section{Simulation}

\begin{figure*}[tbh]
    \centering
    \includegraphics[width=\hsize]{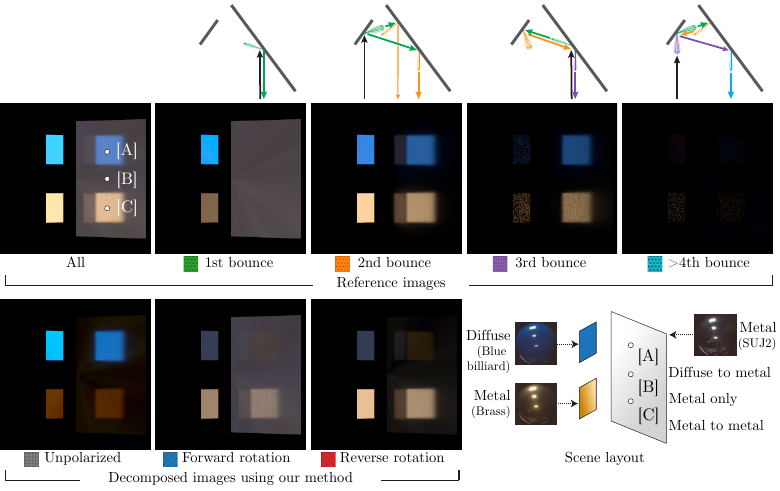}
    \caption{\textbf{Decomposition result using synthetic data.} We designed the scene using multiple planar objects with different materials to produce three types of mixed reflectance components. We can confirm each decomposed image contains the desired reflectance components. Please refer to Fig.~\ref{fig:synthetic_result_quantitative} for quantitative evaluation on the points [A][B][C].}
    \label{fig:synthetic_result}
\end{figure*}

\begin{figure}[h]
    \centering
    \includegraphics[width=\hsize]{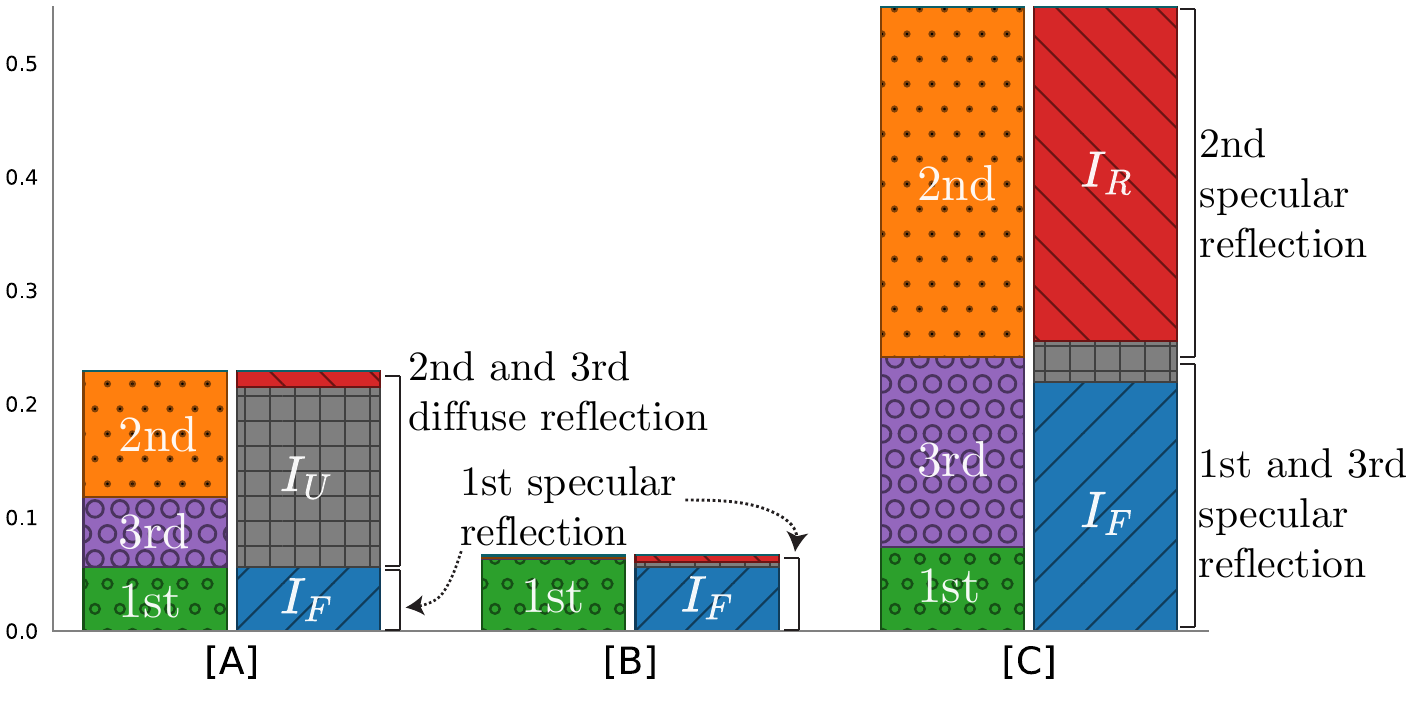}
    \begin{resizebox}{0.999\hsize}{!}{\begin{tabular}{@{}l@{}}
        \begin{tabular}{c|c|cccc|ccc}
            \toprule
                & All & 1st & 2nd & 3rd & {\small$>$}4th & $I_{U}$ & $I_{F}$ & $I_{R}$ \\
            \hline
            {[A]} & 0.2286 & 0.0564 & 0.1106 & 0.0616 & 0.0002 & 0.1592 & 0.0558 & 0.0136 \\
            {[B]} & 0.0664 & 0.0637 & 0.0018 & 0.0009 & 0.0001 & 0.0039 & 0.0563 & 0.0062 \\
            {[C]} & 0.5494 & 0.0734 & 0.3081 & 0.1679 & 0.0000 & 0.0357 & 0.2195 & 0.2942 \\
            \toprule
        \end{tabular}
    \end{tabular}}\end{resizebox}
    
    \caption{\textbf{Quantitative evaluation of decomposed components.} The stacked bar and first table show the intensity values of three points [A][B][C] in Fig.~\ref{fig:synthetic_result}.}
    \label{fig:synthetic_result_quantitative}
\end{figure}

In this section, we evaluate the proposed method using synthetic data. To generate the data, we used Mitsuba~3~\cite{Mitsuba3}, which can simulate the light transport of polarization. We used the measured polarized-BRDF data~\cite{baek2020image} to simulate physically accurate polarization material. We selected \textit{Brass} and \textit{SUJ2} as metal material and \textit{Blue billiard} as diffuse material based on the visual clarity of the rendered image. We arranged the scene with three planar objects in a V-shape so that inter-reflection occurred. The camera and light source are placed close enough together. This light source illuminates the entire scene.

Fig.~\ref{fig:synthetic_result} shows the rendered images as a reference and the decomposed images applied to all mixture light. The top row shows the rendered image (all), as well as individual bounce components (1st, 2nd, 3rd, and more than 4th). Each component is independently rendered by limiting the maximum number of light bounces in Mitsuba~3 renderer. The bottom row shows the decomposed images by using the proposed method. We can confirm each decomposed image contains desired reflection components as explained in Sec.~\ref{sec:method}.

Fig.~\ref{fig:synthetic_result_quantitative} shows the quantitative comparisons of decomposed components at point [A]/[B]/[C]. From this chart, we can observe the following:
\begin{itemize}
     \item Point [A]: This point has a mixture of 1st bounce light from \textit{SUJ2} and 2nd/3rd bounce light from \textit{blue billiard}. The 1st bounce component is specular reflection, and it is corresponded to the forward rotation component ($I_{F}/$1st=0.991). On the other hand, both the 2nd and 3rd bounce component come from diffuse material, and  the sum of intensities is corresponded to the unpolarized component ($I_{U}/$(2nd+3rd)=0.925).
    \item Point [B]: This point is not affected by inter-reflection, almost all the reflected light comes from the 1st bounce, corresponding to the forward rotation component ($I_{F}/$1st=0.884).
    \item Point [C]: This point exhibits a 2nd bounce specular inter-reflection, which corresponds to the reverse rotation components ($I_{R}/$2nd=0.955). Similar to points [A] and [B], the 1st reflection is decomposed as the forward rotation component. There is also a 3rd bounced inter-reflection because the 3rd bounce flips the rotation direction again, becoming the same direction as the 1st ($I_{F}/$(1st+3rd)=0.910).
\end{itemize}

Through this result, we confirm that the decomposed components are matched to the 1st and 2nd components.

\section{Experimental Results}
\begin{figure}[tbh]
    \centering
    \includegraphics[width=0.9\hsize]{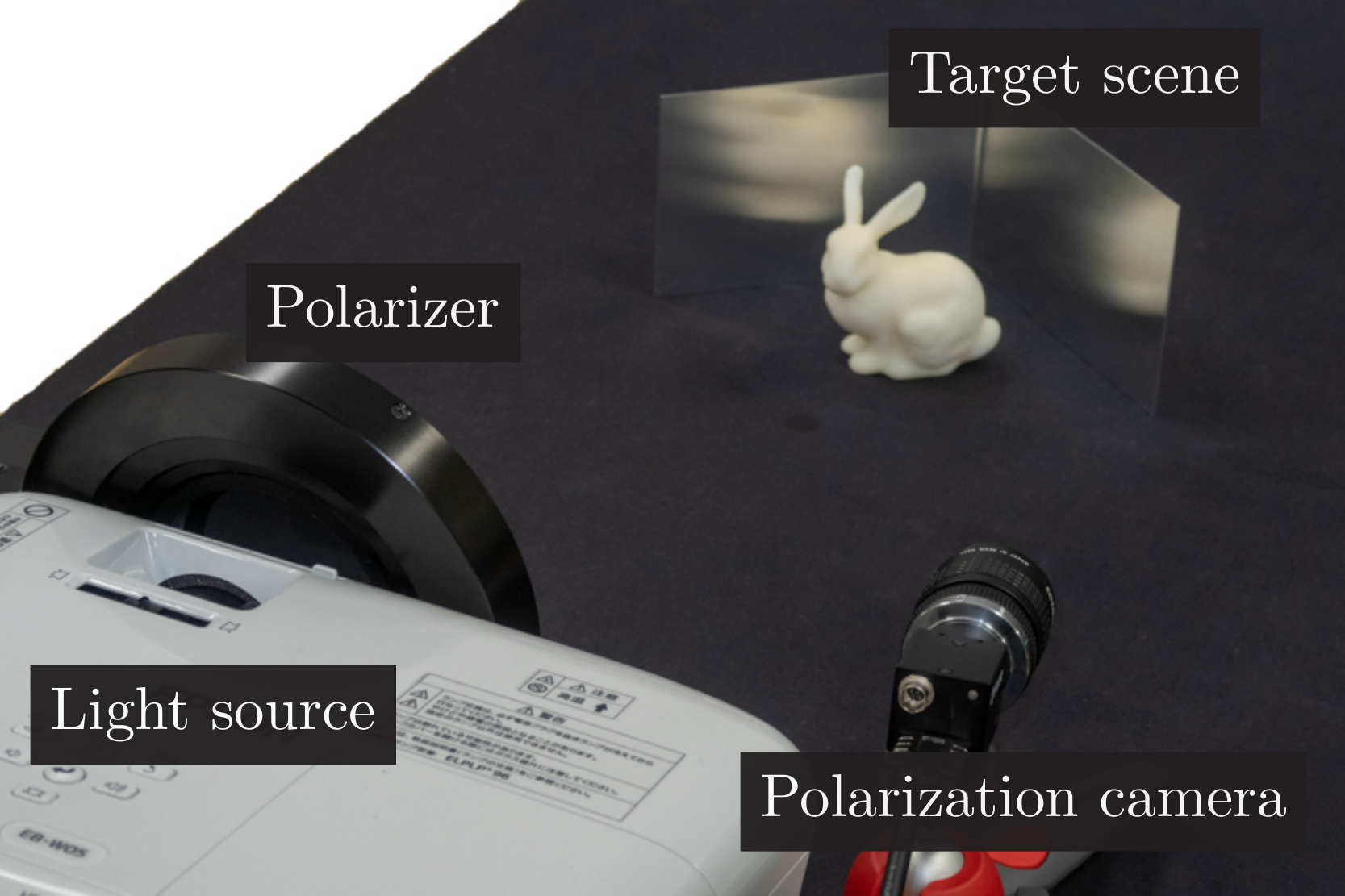}
    \caption{\textbf{Experimental setup for the real scene.} The target scene is illuminated by a light source with a linear polarizer attached to the front and measured by a polarization camera.}
    \label{fig:capture_setup}
\end{figure}

Fig.~\ref{fig:capture_setup} shows the capturing setup for real scenes. The target scene is illuminated by a light source (Epson, EB-W05) with a linear polarizer (Kenko, PRO1D CIRCULAR PL (W)) attached to the front\footnote{We used a 3LCD-type projector, which emits partially linear polarized light, not unpolarized. To mitigate the effect of the polarization, we used a polarizer made up of a quarter waveplate and a linear polarizer. This combination enables us to avoid the influence of polarized light from the projector, and we can illuminate the scene with a constant intensity of linearly polarized light while rotating the polarizer.} and measured by a polarization camera (FLIR, BFS-U3-51S5P-C). This polarization camera can capture the polarization state with a single shot. It is equivalent to a conventional camera with a rotating polarizer. Our method specifically targets specular reflection, which is typically required to capture a high dynamic range (HDR) image. Therefore, we captured sequence images with different exposure times and merged them into a HDR image. We acquire multiple HDR images for every polarizer angle of the light side. As explained in Sec~\ref{subsec:number_of_measurement}, theoretically, the minimum number of images is two, but we captured four images with different polarizer angles on the light source ($\theta_{l}={0,45,90,135}$) in this experiment to be robust to the sensor noise. The measurements were conducted in a darkroom to avoid the environment light. 

\subsection{Decomposition of Reflection Components}
\begin{figure*}[tbh]
    \centering
    \begin{minipage}{0.2\hsize}
        \centering
        \includegraphics[width=0.98\hsize]{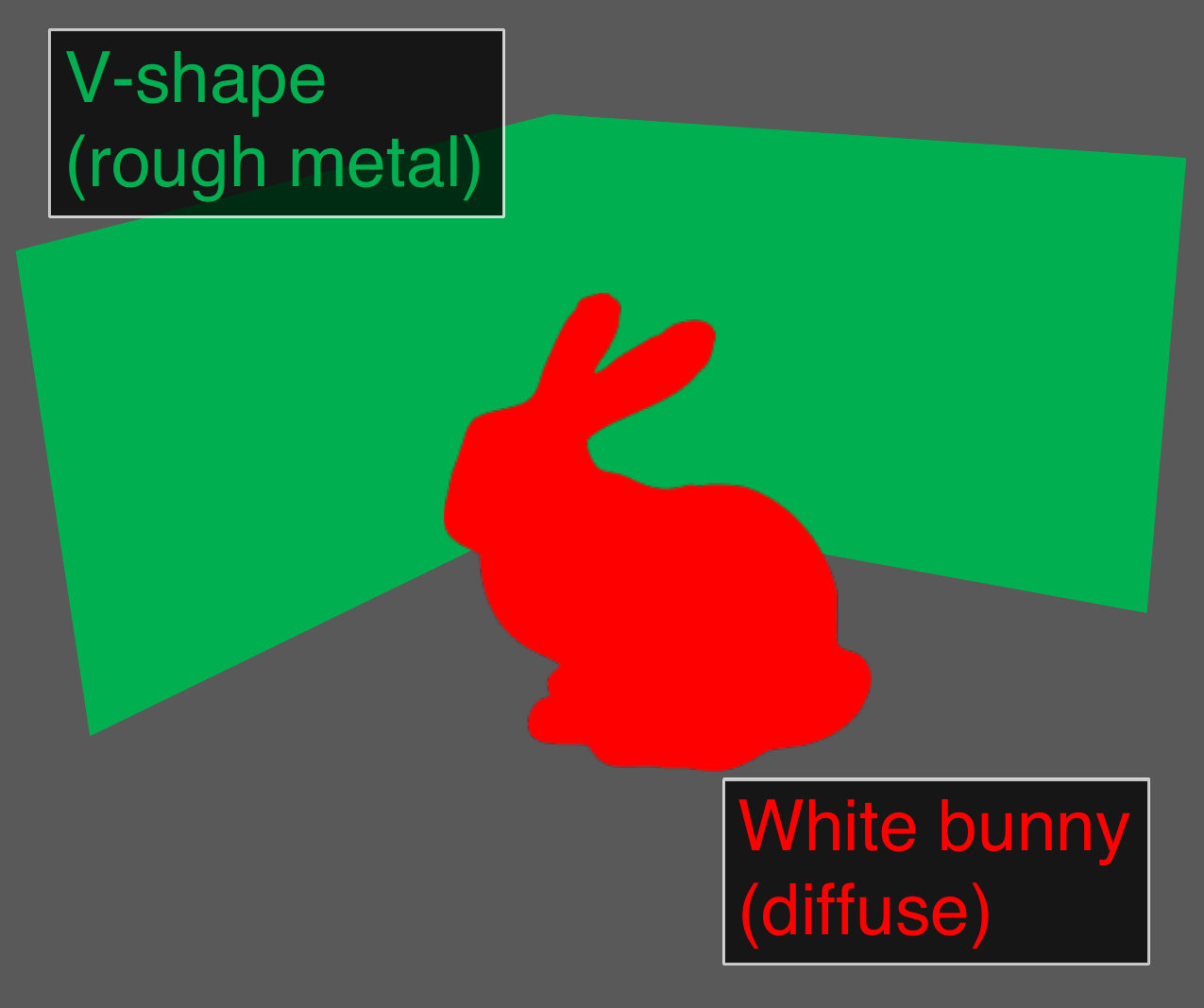}
    \end{minipage}%
    \begin{minipage}{0.2\hsize}
        \centering
        \includegraphics[width=0.98\hsize]{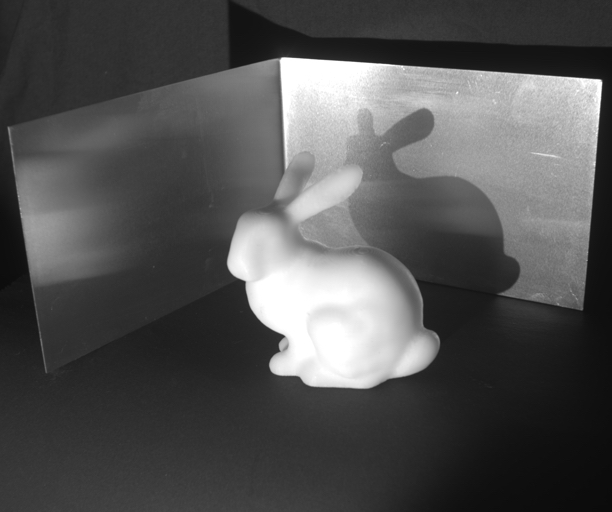}
    \end{minipage}%
    \begin{minipage}{0.2\hsize}
        \centering
        \includegraphics[width=0.98\hsize]{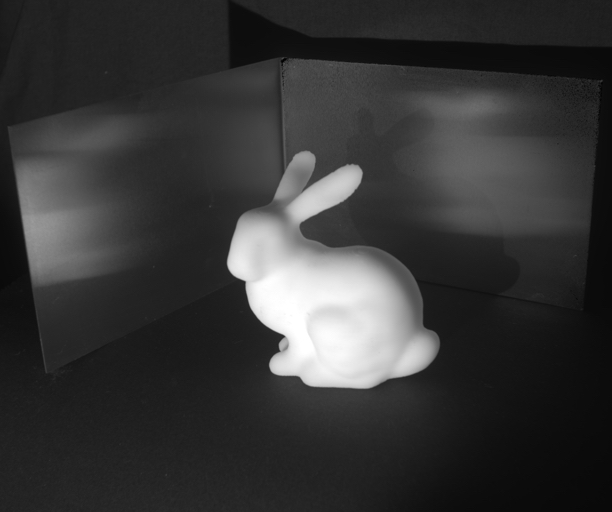}
    \end{minipage}%
    \begin{minipage}{0.2\hsize}
        \centering
        \includegraphics[width=0.98\hsize]{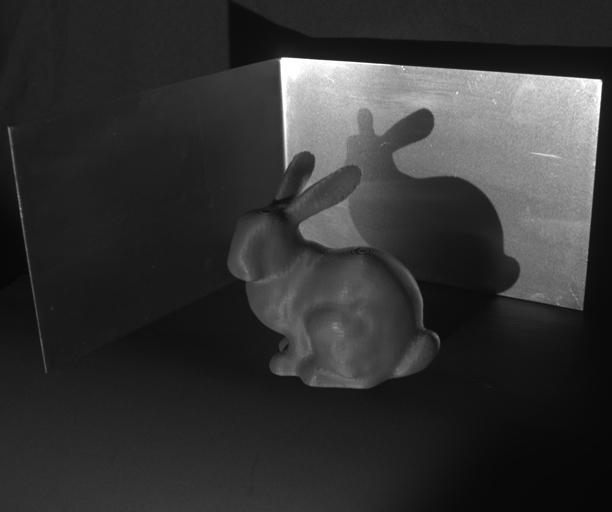}
    \end{minipage}%
    \begin{minipage}{0.2\hsize}
        \centering
        \includegraphics[width=0.98\hsize]{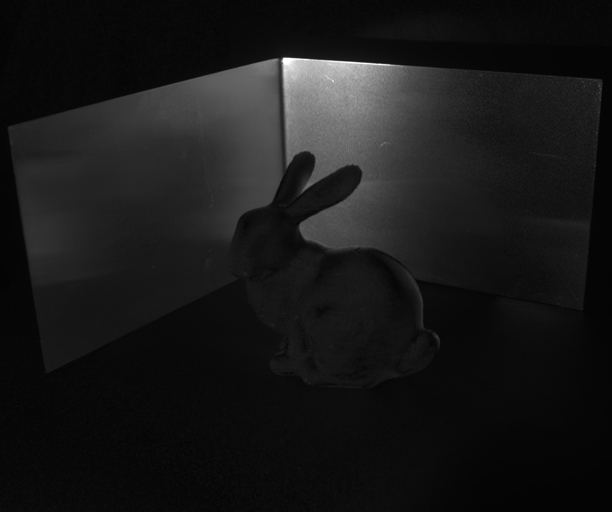}
    \end{minipage}\\  
    \vspace{0.3ex}
    \begin{minipage}{0.2\hsize}
        \centering
        \includegraphics[width=0.98\hsize]{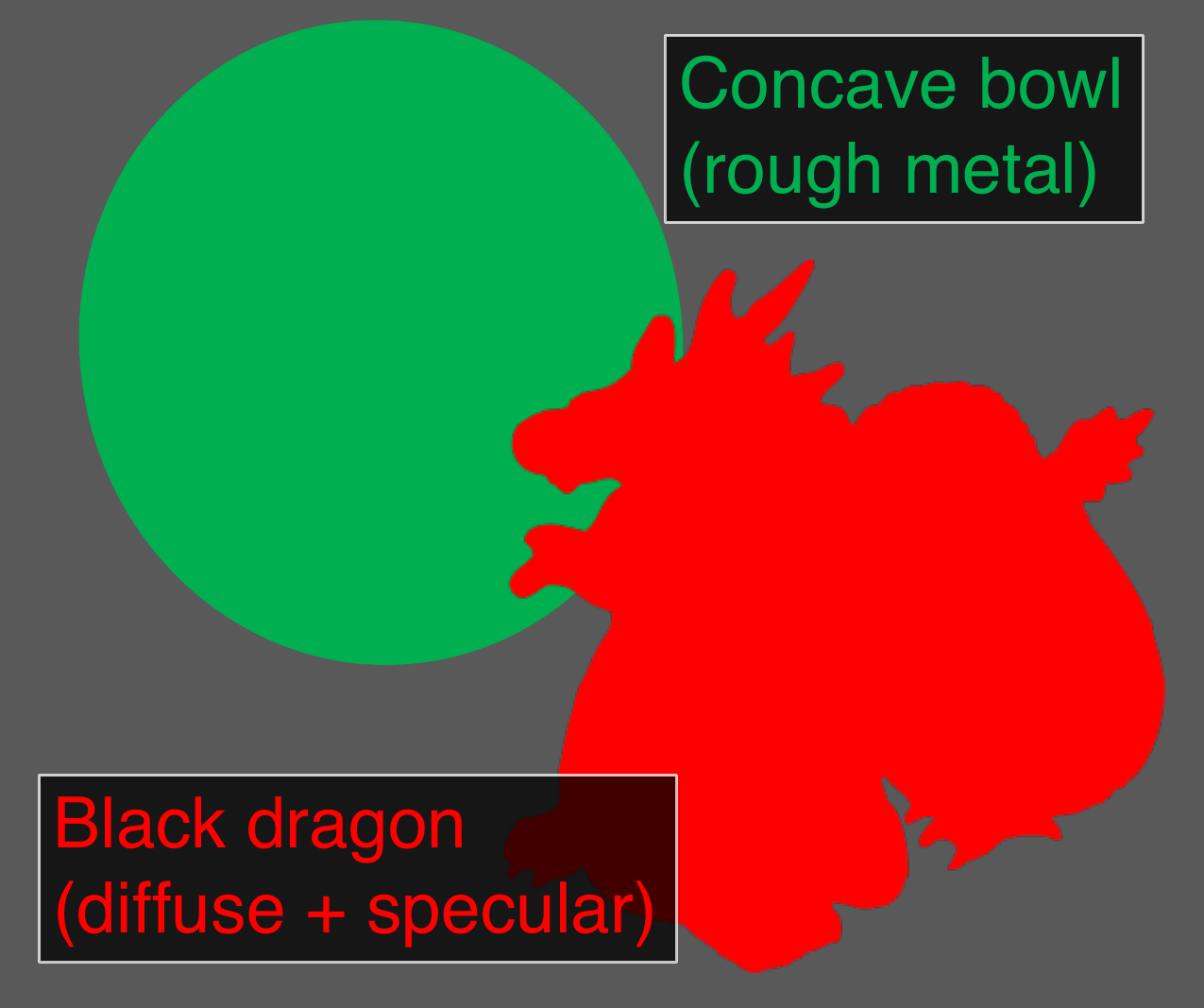}
    \end{minipage}%
    \begin{minipage}{0.2\hsize}
        \centering
        \includegraphics[width=0.98\hsize]{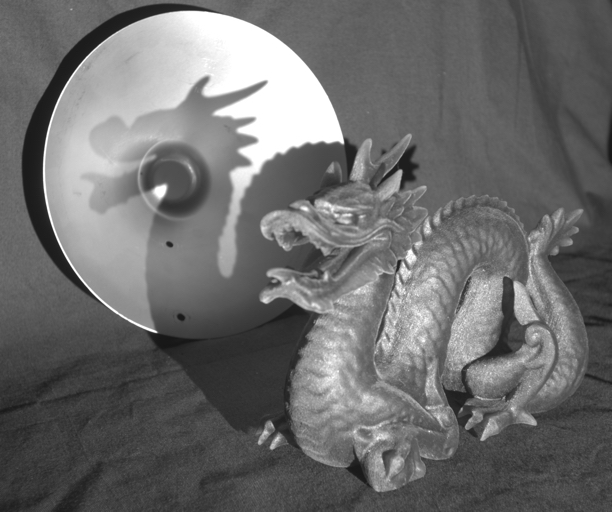}
    \end{minipage}%
    \begin{minipage}{0.2\hsize}
        \centering
        \includegraphics[width=0.98\hsize]{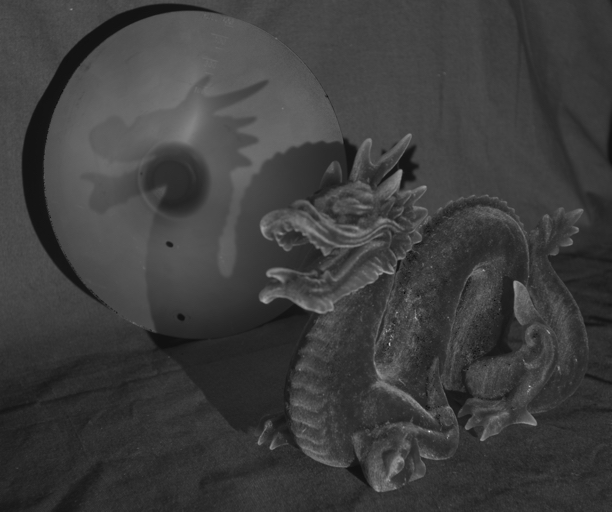}
    \end{minipage}%
    \begin{minipage}{0.2\hsize}
        \centering
        \includegraphics[width=0.98\hsize]{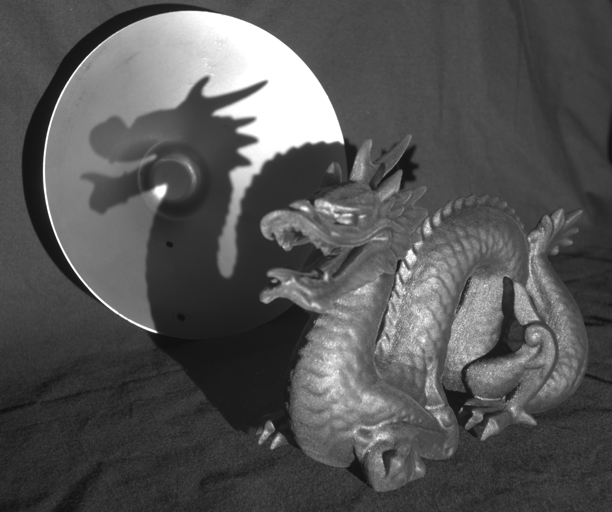}
    \end{minipage}%
    \begin{minipage}{0.2\hsize}
        \centering
        \includegraphics[width=0.98\hsize]{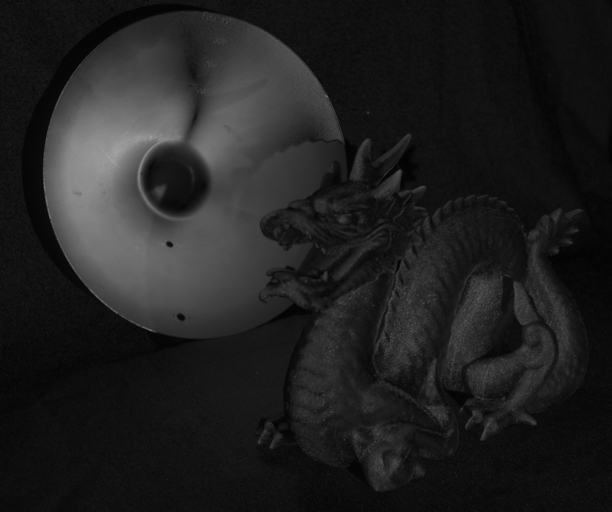}
    \end{minipage}\\  
    \vspace{0.3ex}
    \begin{minipage}{0.2\hsize}
        \centering
        \includegraphics[width=0.98\hsize]{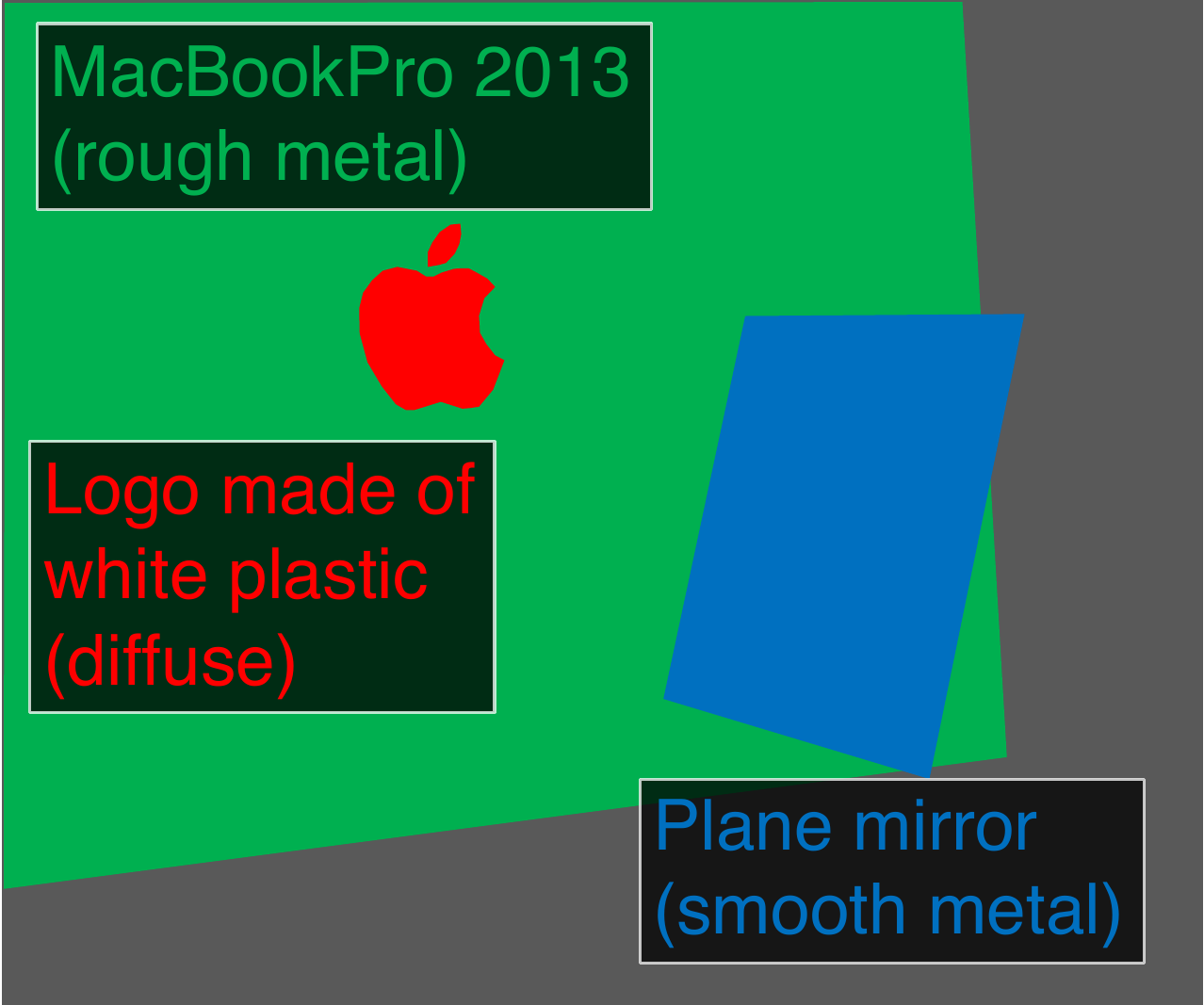}
    \end{minipage}%
    \begin{minipage}{0.2\hsize}
        \centering
        \includegraphics[width=0.98\hsize]{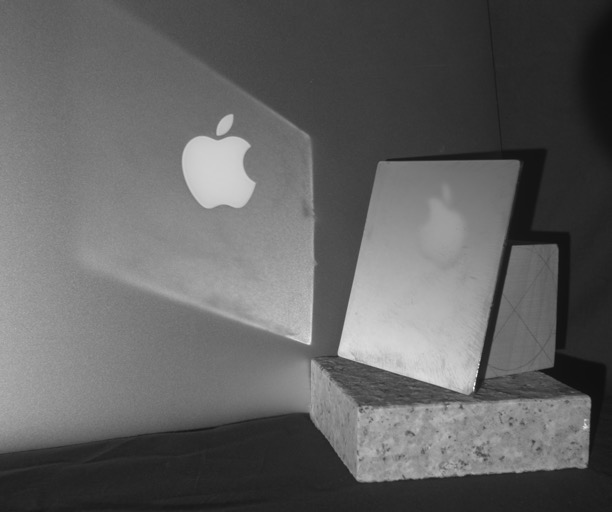}
    \end{minipage}%
    \begin{minipage}{0.2\hsize}
        \centering
        \includegraphics[width=0.98\hsize]{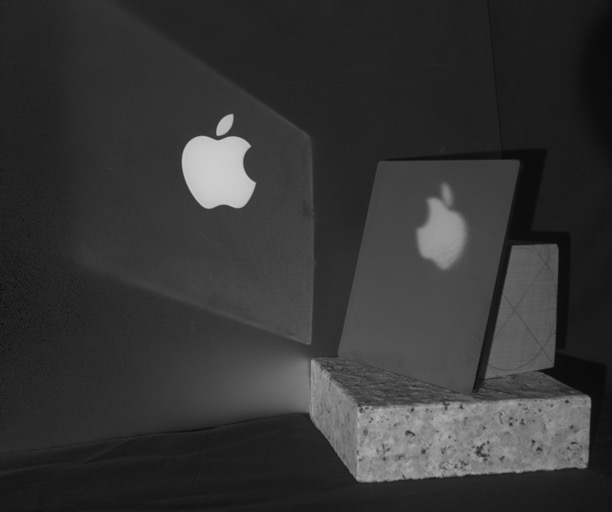}
    \end{minipage}%
    \begin{minipage}{0.2\hsize}
        \centering
        \includegraphics[width=0.98\hsize]{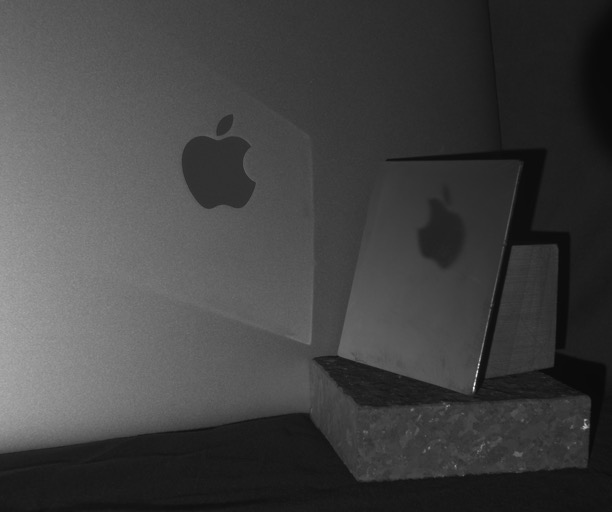}
    \end{minipage}%
    \begin{minipage}{0.2\hsize}
        \centering
        \includegraphics[width=0.98\hsize]{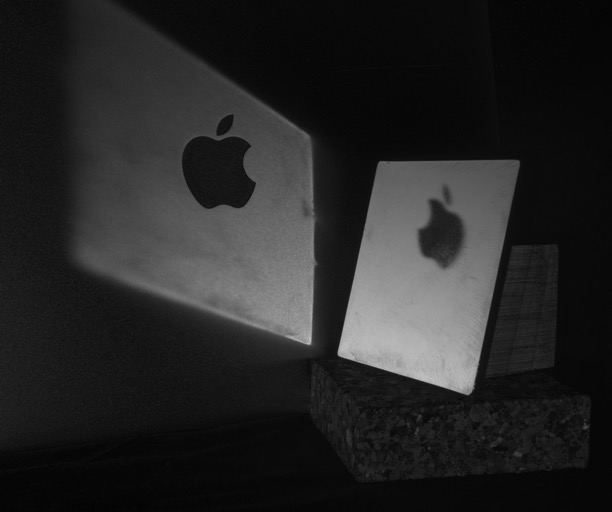}
    \end{minipage}\\  
    \vspace{0.3ex}
    \begin{minipage}{0.2\hsize}
        \centering
        \includegraphics[width=0.98\hsize]{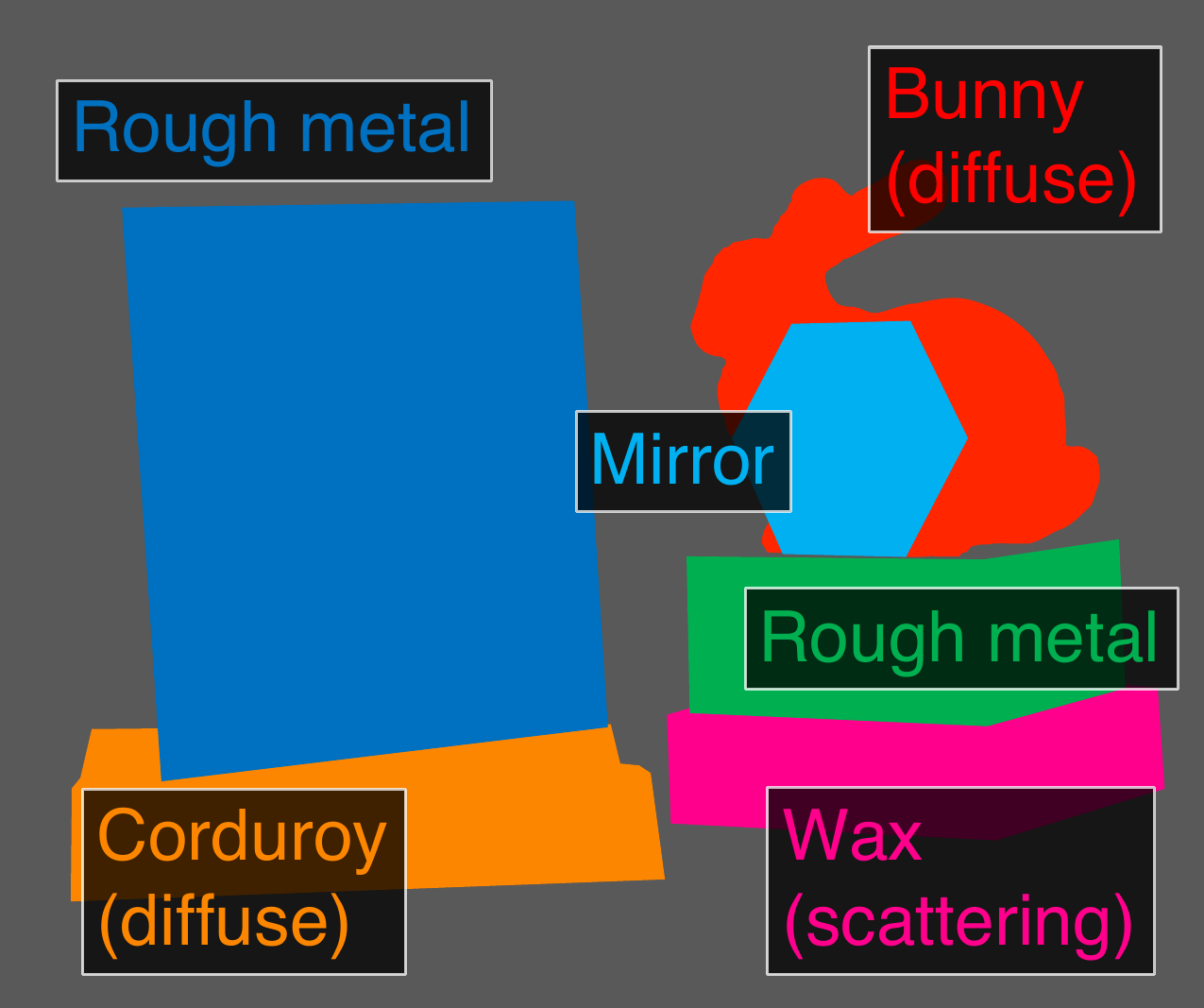}
    \end{minipage}%
    \begin{minipage}{0.2\hsize}
        \centering
        \includegraphics[width=0.98\hsize]{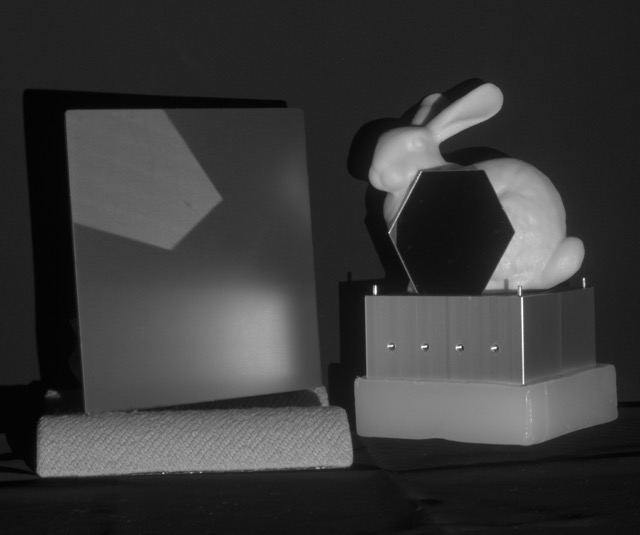}
    \end{minipage}%
    \begin{minipage}{0.2\hsize}
        \centering
        \includegraphics[width=0.98\hsize]{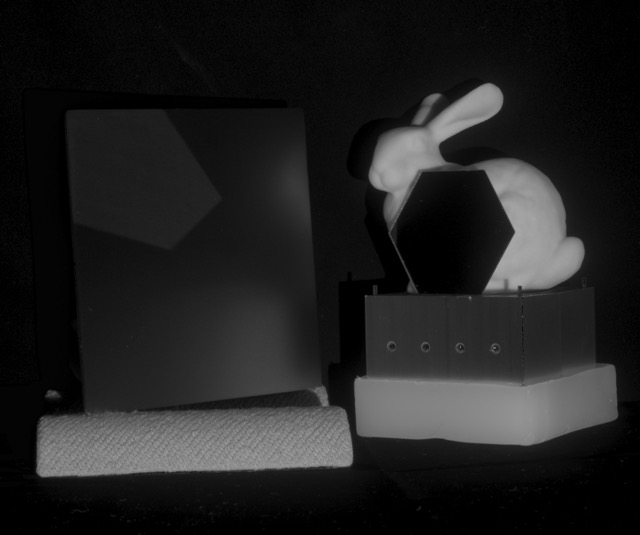}
    \end{minipage}%
    \begin{minipage}{0.2\hsize}
        \centering
        \includegraphics[width=0.98\hsize]{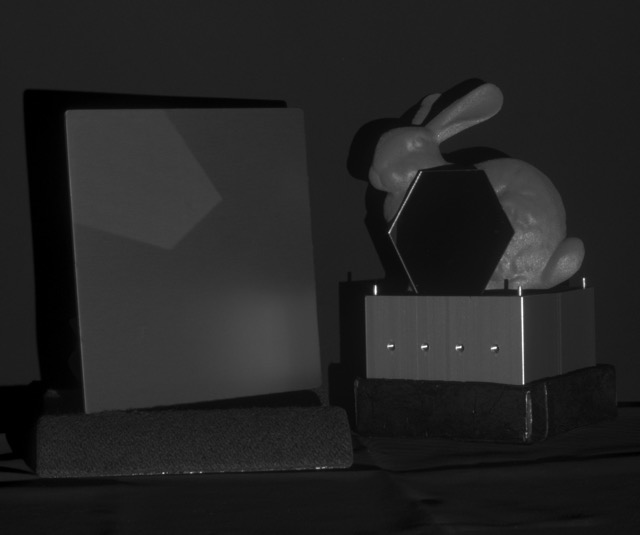}
    \end{minipage}%
    \begin{minipage}{0.2\hsize}
        \centering
        \includegraphics[width=0.98\hsize]{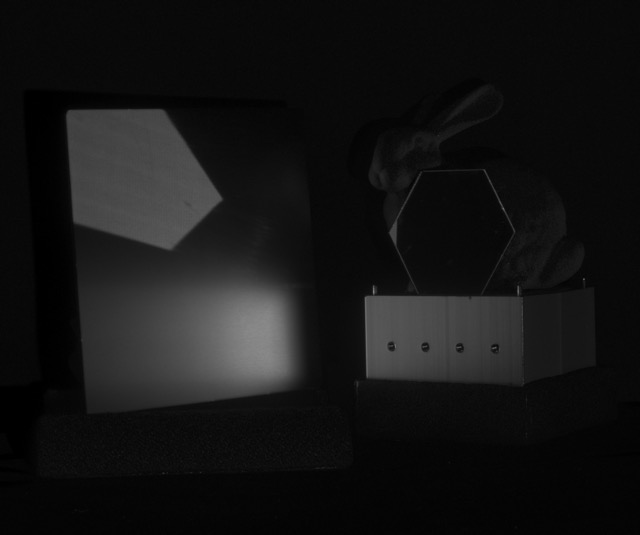}
    \end{minipage}\\  
    \vspace{1ex}
    \myunderbracket[1ex]{
    \begin{minipage}[t]{0.2\hsize}
        \centering
        \phantom{Phantom p}
    \end{minipage}}{Scene layout}%
    \myunderbracket[1ex]{
    \begin{minipage}[t]{0.2\hsize}
        \centering
        \phantom{Phantom p}
    \end{minipage}}{Regular (all)}%
    \myunderbracket[1ex]{
    \begin{minipage}[t]{0.2\hsize}
        \centering
        Unpolarized $I_{U}$
    \end{minipage}%
    \begin{minipage}[t]{0.2\hsize}
        \centering
        Forward rotation $I_{F}$
    \end{minipage}%
    \begin{minipage}[t]{0.2\hsize}
        \centering
        Reverse rotation $I_{R}$
    \end{minipage}}{Decomposition result}
    \caption{\textbf{Decomposition results for various scenes.} Each column shows the images of the scene layout, regular (sum of all components), unpolarized, forward rotation, and reverse rotation components. Our method can extract specular inter-reflection of metal objects, as shown in the reverse rotation component for various scenes.}
    \label{fig:result_decomposition}
\end{figure*}

Fig.~\ref{fig:result_decomposition} shows the decomposed result in the real scenes. Our method successfully decomposes the desired reflectance components for various scenes. The reverse components have a majority of specular inter-reflection for various geometry and metallic materials.

\subsection{Comparison with high-frequency illumination method}
\begin{figure*}[tbh]
    \centering
    \begin{minipage}[t]{0.25\hsize}
        \centering
        \includegraphics[width=0.99\hsize]{figures/nayar/intensity_all.jpeg}
        Regular (all)
    \end{minipage}%
    \begin{minipage}[t]{0.25\hsize}
        \centering
        \begin{tikzpicture}
            \definecolor{high_specular_blue}{RGB}{101, 120, 180}
            \node[above right, inner sep=0] (image) at (0,0) { \includegraphics[width=0.99\hsize]{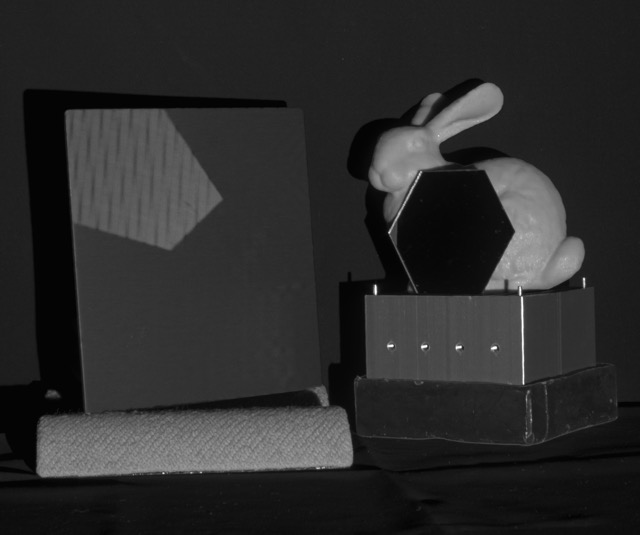} };
            \begin{scope}[x={($0.1*(image.south east)$)}, y={($0.1*(image.north west)$)}]                
                \draw[thick, high_specular_blue, dashed] (1.0, 7.9) -- (1.15, 5.85) -- (2.7, 5.35) -- (3.5, 6.33) -- (2.55, 7.96) -- cycle;
                \draw[latex-, thick, high_specular_blue, opacity=0.9] (3, 7.3) -- ++(0.2, 1.2) node[align=left, anchor=west, text=high_specular_blue, text opacity=1.0, fill=black, fill opacity=0.8]{\footnotesize High-frequency \\\footnotesize specular inter-reflection};
            \end{scope}        
        \end{tikzpicture}
        Direct
    \end{minipage}%
    \begin{minipage}[t]{0.25\hsize}
        \centering
        \begin{tikzpicture}
            \definecolor{low_specular_teal}{RGB}{0, 151, 157}
            \node[above right, inner sep=0] (image) at (0,0) { \includegraphics[width=0.99\hsize]{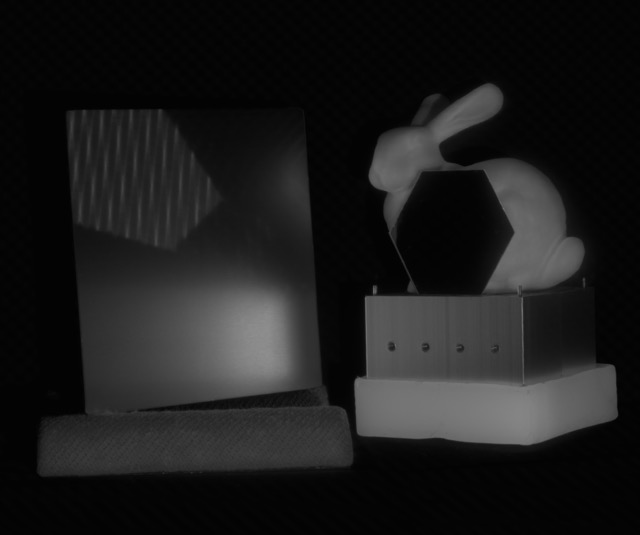} };
            \begin{scope}[x={($0.1*(image.south east)$)}, y={($0.1*(image.north west)$)}]
                \draw[thick, low_specular_teal, dashed] (2.3, 2.9) -- (4.7, 2.9) -- (4.6, 4.7) -- (2.3, 4.7) --cycle;
                \draw[latex-, thick, low_specular_teal, opacity=0.9] (4.6, 4) -- (6.2, 2.5) node[align=left, anchor=north, text=low_specular_teal, text opacity=1.0, fill=black, fill opacity=0.8]{\footnotesize Low-frequency \\\footnotesize specular inter-reflection};
            \end{scope}        
        \end{tikzpicture}
        Global
    \end{minipage}%
    \begin{minipage}[t]{0.25\hsize}
        \centering
        \begin{tikzpicture}
            \definecolor{high_specular_blue}{RGB}{101, 120, 180}
            \definecolor{low_specular_teal}{RGB}{0, 151, 157}
            \node[above right, inner sep=0] (image) at (0,0) { \includegraphics[width=0.99\hsize]{figures/nayar/intensity_reverse.jpeg} };
            \begin{scope}[x={($0.1*(image.south east)$)}, y={($0.1*(image.north west)$)}]
                \draw[thick, high_specular_blue, dashed] (1.0, 7.9) -- (1.15, 5.85) -- (2.7, 5.35) -- (3.5, 6.33) -- (2.55, 7.96) -- cycle;
                \draw[thick, low_specular_teal, dashed] (2.3, 2.9) -- (4.7, 2.9) -- (4.6, 4.7) -- (2.3, 4.7) --cycle;
            \end{scope}        
        \end{tikzpicture}
        Reverse rotation $I_{R}$
    \end{minipage}\\ 
    \begin{minipage}[t]{0.25\hsize}
        \centering
        \hspace*{\hsize}
    \end{minipage}%
    \begin{minipage}[t]{0.5\hsize}
        \centering
        \myunderbracket{\hspace*{\hsize}}{Nayar\etal~\cite{nayar2006fast}}
    \end{minipage}%
    \begin{minipage}[t]{0.25\hsize}
        \centering
        \myunderbracket{\hspace*{\hsize}}{Ours}
    \end{minipage}
    \caption{\textbf{Comparison with Nayar's high-frequency illumination method for specular inter-reflection.} This scene contains two types of specular inter-reflection categorized as global components but have distinct spatial frequency responses. Nayar's method failed to separate high-frequency inter-reflection from the hexagonal mirror. In contrast, our method extracts both specular inter-reflection regardless of its frequency response.}
    \label{fig:result_vs_nayar}
\end{figure*}

Compared to the other decomposition methods, our method can analyze specular inter-reflection regardless of the spatial frequency response. We compare our method with Nayar’s high-frequency illumination method~\cite{nayar2006fast}. Their method can decompose direct and global (including inter-reflection) light components. However, as their paper mentioned, their method fails to decompose high-frequency specular inter-reflection. For instance, specular inter-reflection from a perfect mirror is difficult to distinguish from direct reflection. Fig.~\ref{fig:result_vs_nayar} shows the comparison of decomposition results. We can confirm that Nayar’s method fails to decompose high-frequency specular inter-reflection from the hexagon-shaped mirror. In contrast, our method successfully extracts specular inter-reflection.

\subsection{Combination with high-frequency illumination method}

We show the combination of the different light transport analysis methods to achieve a detailed decomposition. Each cue has different limitations for targeting optical phenomena. For example, our polarization-based method is incapable of distinguishing diffuse reflection and subsurface scattering; on the other hand, Nayar's method fails for high-frequency inter-reflection cases. Therefore, the capability of a combination of complementary methods is crucial to achieving a detailed analysis of light transport.
Our method utilizes polarization as a cue, which does not cause any conflicts with other cues like spatial-frequency and time-of-flight based methods. 

\begin{figure*}[tbh]
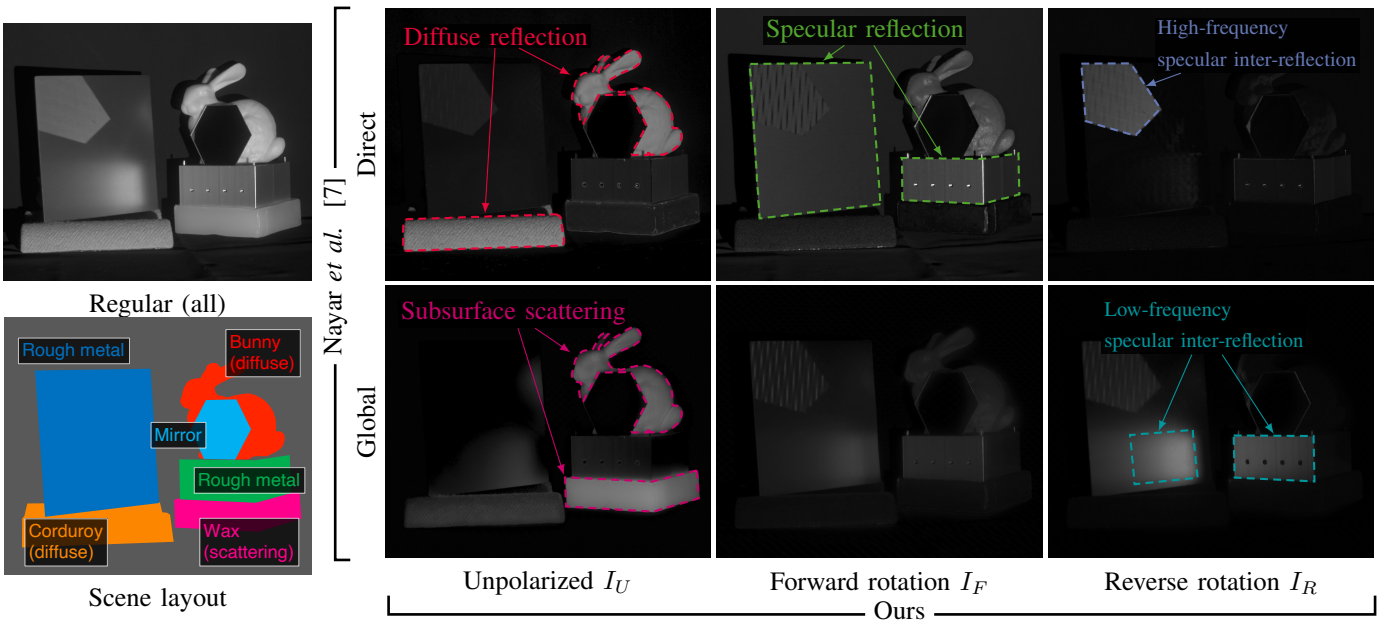

    \centering
    \begin{minipage}[c]{0.23\hsize}
        \begin{minipage}[c]{\hsize}
            \centering
            \includegraphics[width=0.98\hsize]{figures/nayar/intensity_all.jpeg}
            Regular (all)
        \end{minipage}\\
        \begin{minipage}[c]{\hsize}
            \centering
            \includegraphics[width=0.98\hsize]{figures/nayar/setup.pdf}
            Scene layout
        \end{minipage}
    \end{minipage}%
    \begin{minipage}[c]{0.76\hsize}
        \centering
        \input{figures/nayar/table.tex}
    \end{minipage}
    \caption{\textbf{Combining with Nayar's decomposition method.} \textit{Row:} Direct / Global components with Nayar's method~\cite{nayar2006fast}. \textit{Column:} Unpolarized / Forward / Reverse components (ours). As the two methods differ in targeting optical phenomena, we can combine them in a complementary manner and achieve a more in-depth decomposition. In particular, this combination extracts only the high-frequency or low-frequency specular inter-reflections that have never been achieved by previous methods.}
    \label{fig:result_with_nayar}
\end{figure*}

Fig.~\ref{fig:result_with_nayar} shows the results of the combination of Nayar's method~\cite{nayar2006fast}. Their method separates direct and global components by projecting multiple high-frequency checker patterns from a projector. To combine with their method, we first apply Nayar's method and obtain direct and global components for each polarization state. Then, we apply our method for direct and global components, respectively. Even with this straightforward combination, they work complementary to each other and achieve detailed light transport analysis.

\subsection{3D Measurement for Specular Inter-reflection}
As an application, we utilize our decomposition method for 3D measurement with the projector-camera system. Active lighting methods (e.g., structured-light, photometric stereo, and time-of-flight), including the projector-camera system, sometimes struggle with the inter-reflection. This is because these methods assume only direct reflection, which can be problematic when measuring metallic objects with strong specular inter-reflection. Our decomposition method reduces the specular inter-reflection and enables robust measurement of metal objects. 

\begin{figure*}[tbh]
    \centering
    \includegraphics[width=\hsize]{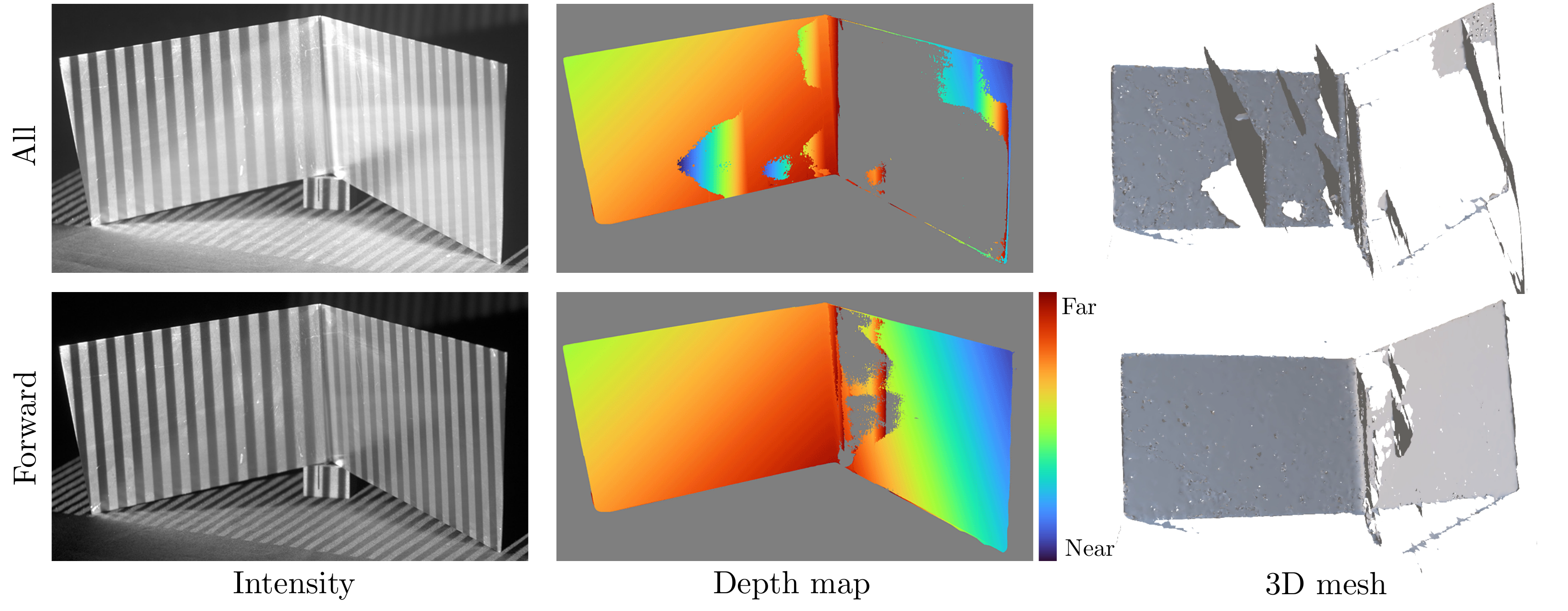}
    
    \caption{\textbf{Application for 3D measurement under strong specular inter-reflection.} We measure the V-groove metal object using Gray code structured light. \textit{Left:} An example of an intensity image projected under a Gray code pattern. \textit{Middle:} Reconstructed depth map. The gray color areas in the image indicate that they are either out of range of depth or background. \textit{Right:} Reconstructed 3D mesh from a different viewpoint. Without our method (all), due to the specular inter-reflection, the proportion of correct points is only 54.6\%. With our method, by using the forward rotation components, the proportion of correct points increases to 92.8\%.}
    \label{fig:result_3d_measurement}
\end{figure*}

Fig.~\ref{fig:result_3d_measurement} shows the results of 3D measurement with and without our method. In this experiment, we use the Gray code pattern as structured light. To reduce the specular inter-reflection, we used the forward rotation intensity to estimate 3D points. 
For the quantitative evaluation, we used the proportion of the correct 3D points. We fitted two planes of a V-groove on measured 3D points and used them as ground truth. Then, we counted the 3D points within 1mm of a ground truth plane. From a number of points, we calculated the proportion as the correct points.
In the traditional method without decomposition, inter-reflection causes errors in most points. On the other hand, using the forward rotation component expands the region of the correct points.

\section{Discussion}
\subsection{Why Not Circular Polarization?}
We proposed a measurement method by rotating the plane of linearly polarized light in order to decompose the reflected components. We clarify why we use linear polarization instead of using circular polarization. Similar to the case of linear polarization discussed in Sec.\ref{sec:principle}, the left- or right-hand circular polarization flips between single and double bounce reflection~\cite{baek2021polarimetric}. We can utilize circular polarization to distinguish direct and inter-reflection, as long as they are not mixed. However, it becomes impossible to decompose if both reflection components are mixed because the incoherent mixture of right-hand and left-hand circular polarized light is observed as unpolarized\footnote{Mathematically, polarized light can be described by Stokes parameters. The incoherent mixture of left-hand and right-hand circularly polarized lights $S_\textrm{left} = (1, 0, 0, -1)$, $S_\textrm{right} = (1, 0, 0, 1)$ is a summation of them $S_\textrm{left}+S_\textrm{right}=(1, 0, 0, 0)$ which is same to unpolarized light.}. That is why we can not use circular polarization alone for this purpose.

\subsection{Phase of AoLP}
\begin{figure}[tbh]
    \centering
    \begin{minipage}{0.309\hsize}
        \centering
        \includegraphics[width=0.99\hsize]{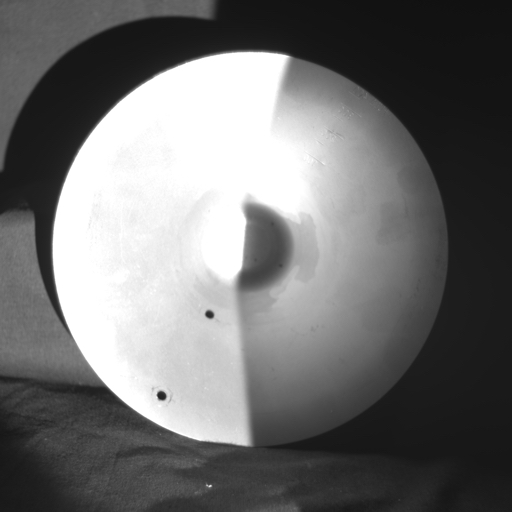}
        {\small Intensity}
    \end{minipage}%
    \begin{minipage}{0.309\hsize}
        \centering
        \includegraphics[width=0.99\hsize]{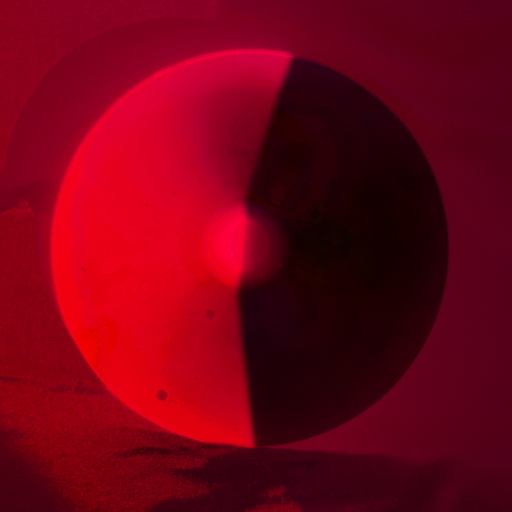}
        {\small Phase (forward) $\phi_{F}$}
    \end{minipage}%
    \begin{minipage}{0.309\hsize}
        \centering
        \includegraphics[width=0.99\hsize]{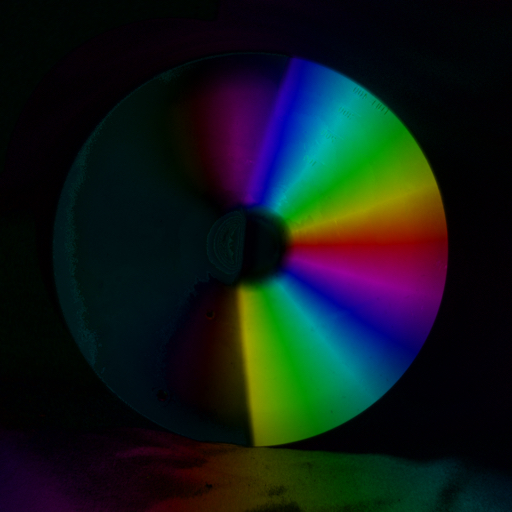}
        {\small Phase (reverse) $\phi_{R}$}
    \end{minipage}%
    \begin{minipage}{0.074\hsize}
        \centering
        \includegraphics[width=0.99\hsize]{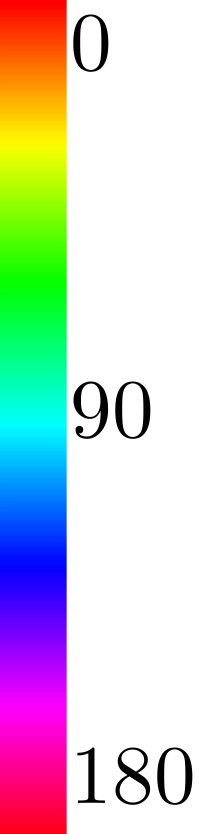}
        {\small \vphantom{Intensity}}
    \end{minipage}
    
    \caption{\textbf{Visualization of phase components.} The scene setup is identical to Fig.~\ref{fig:verify_rotation}. The phase values are colored by mapping them to the Hue circle, and the Brightness is modulated by the decomposed intensities.}
    \label{fig:result_phase}
\end{figure}

Since the experimental results only showed the intensity images, here we show and describe the phase ($\phi_F$ and $\phi_R$) obtained from our decomposition method. Fig.~\ref{fig:result_phase} shows the phase image of forward and reverse rotation components. 
The phase of the forward component is near zero for all pixels because direct reflection does not change the polarization phase. We can also see this fact from the plot of direct reflection shown in Fig. \ref{fig:verify_rotation}(b), where the intercept is zero.
In contrast, the phase of the reverse component varies depending on the direction from where the first bounce light is coming. For instance, the horizontal and vertical directions do not cause a phase shift, while the oblique directions result in a phase shift. This fact can also be seen from the [C] in Fig.\ref{fig:verify_rotation}(b). Currently, we do not use this phase information, but it can potentially analyze the scene geometry.

\section{Limitation}
\label{sec:limitation}
\subsection{Fresnel Reflection}
Our method has a limitation for Fresnel reflection of dielectric material as it is not explicitly incorporated in the model. Our model approximates the reflectance value as a constant with respect to the rotation of the polarizer on the light source side. However, in Fresnel reflection, the reflectance varies between s and p polarization. Consequently, the intensity of reflected light changes while rotating the polarizer on the light source side. This effect becomes dominant when the diffuse reflection is relatively small, such as black plastic.

\subsection{3rd Bounce Light}
Our method also has a limitation for multiple bounce inter-reflection (more than 3rd). The direction of polarization rotation flips upon reflection, causing the 3rd bounce to have the same rotation as the 1st. Therefore, our method cannot distinguish 1st and 3rd bounce reflections, and they would be decomposed as forward rotation components. 
This is the nature of polarization behavior, and resolving this problem is fundamentally difficult with polarization alone. Generally, the 3rd bounce light tends to be weaker than the 1st/2nd bounce with rough metal surfaces. However, some certain geometry and materials cause strong 3rd bounce reflection and become problematic. The result shown in Fig.~\ref{fig:synthetic_result} is the exact case where the 3rd component is intentionally emphasized because the smooth specular surfaces preserve the intensity of reflected light.

Despite this limitation, our method is still valuable. In light transport analysis, we can use other optical cues along with polarization. Previous works have already shown promising results by combining polarization with other cues such as color~\cite{nayar1997separation}, high-frequency pattern~\cite{takatani2018decomposition}, and Time-of-Flight~\cite{baek2021polarimetric}. Our method contributes to these previous and future methods to enable a comprehensive analysis of light transport, including higher-order multiple bounce reflection. Furthermore, our method alone can remove undesirable 2nd specular inter-reflection. For example, as already shown in the result, our method is effective for 3D imaging even though the scene contains 3rd bounce light.
\section{Conclusion}
In this study, we proposed a novel decomposition method of reflection components by featuring the rotation direction of the polarization plane. Our proposed method enables us to decompose the specular inter-reflections, which had been difficult to handle in the past. We verified the effectiveness of our method in both synthetic and real data. Our method can be combined with other methods to explore a detailed analysis of light transport. As an application of 3D measurement, our method reduced the error that occurred by strong specular inter-reflection.

\bibliographystyle{IEEEtran}
\bibliography{IEEEabrv,references}
{
\appendix[Solving the Decomposition Equation]


This appendix describes the transformation to solve Eq.~\ref{equ:non-linear_optimization_problem}.
At the beginning, we convert Eq.~\ref{equ:sum_of_three_components}
into linear form,
\begin{equation}
    I(\theta_{c}, \theta_{l}) = \mathbf{w}(\theta_{c}, \theta_{l}) \mathbf{X},
    \label{equ:I-matrix}
\end{equation}
where matrices of known values $W$ and unknown parameters $\mathbf{X}$ are described as follows.
\begin{equation}
    \mathbf{w}(\theta_{c}, \theta_{l}) = 
    \begin{bmatrix}
        1 \\ 
        \cos{2\theta_{c}}\cos{2\theta_{l}} \\ \sin{2\theta_{c}}\sin{2\theta_{l}} \\
        \cos{2\theta_{c}}\sin{2\theta_{l}} \\ \sin{2\theta_{c}}\cos{2\theta_{l}}  
    \end{bmatrix},
\end{equation}
\begin{equation}
    \mathbf{X} = 
    \begin{bmatrix}
    x_{1} \\ x_{2} \\ x_{3} \\ x_{4} \\ x_{5}
    \end{bmatrix}
    = \frac{1}{2}
    \begin{bmatrix}
    I_{U}+I_{F}+I_{R} \\
    I_{F}\cos{2\phi_{F}}+I_{R}\cos{2\phi_{R}} \\
    -I_{F}\cos{2\phi_{F}}+I_{R}\cos{2\phi_{R}} \\ 
    I_{F}\sin{2\phi_{F}}-I_{R}\sin{2\phi_{R}} \\
    I_{F}\sin{2\phi_{F}}+I_{R}\sin{2\phi_{R}}
    \end{bmatrix}.
\end{equation}
Suppose we capture $N$ frames while rotating the polarizers of camera ($\theta_{c}=\theta_{c_{1}},\theta_{c_{2}},\cdots,\theta_{c_{N}}$) and light source ($\theta_{l}=\theta_{l_{1}},\theta_{l_{2}},\cdots,\theta_{l_{N}}$), and get a sequence of intensities $I(\theta_{c_k}, \theta_{l_k})$ and vectors $\mathbf{w}(\theta_{c_k}, \theta_{l_k})$. We can stack them as the column vector $\mathbf{I}$ and the matrix $\mathbf{W}$ as follows
\begin{equation}
    \mathbf{I}=
    \begin{bmatrix}
    I(\theta_{c_{1}}, \theta_{l_{1}}) \\
    I(\theta_{c_{2}}, \theta_{l_{2}}) \\
    \vdots \\
    I(\theta_{c_{N}}, \theta_{l_{N}})
    \end{bmatrix}, 
    \mathbf{W}=
    \begin{bmatrix}
    - & \mathbf{w}(\theta_{c_{1}}, \theta_{l_{1}}) & - \\
    - & \mathbf{w}(\theta_{c_{2}}, \theta_{l_{2}}) & - \\
    & \vdots & \\
    - & \mathbf{w}(\theta_{c_{N}}, \theta_{l_{N}}) & -
    \end{bmatrix}.
\end{equation}
and we can extend Eq.~\ref{equ:I-matrix} to
\begin{equation}
    \label{equ:I=WX}
    \mathbf{I}=\mathbf{W}\mathbf{X}.
\end{equation}
To solve the above equation, the matrix $\mathbf{W}$ must have its own inverse matrix. Fortunately, if we choose the combination of $(\theta_{c}, \theta_{l})$ carefully, $\mathbf{W}$ becomes invertible (we discuss it later). 
If we have more than five observations ($N\geq5$), we can use the pseudo-inverse matrix of $\mathbf{W}$ to be robust to the sensor noise. With the pseudo-inverse matrix $\mathbf{W}^{+}$, we can solve $\mathbf{X}$ as follows
\begin{equation}
    \label{equ:calc-X}
    \begin{split}
        \mathbf{X} &= (\mathbf{W}^{T}\mathbf{W})^{-1}\mathbf{W}^{T} \mathbf{I} \\
               &= \mathbf{W}^{+} \mathbf{I}.
    \end{split}
\end{equation}
By using solved $\mathbf{X}$, we can calculate the parameters of each reflection component $I_{U},I_{F},\phi_{F},I_{R},\phi_{R}$. The relationships among $x_{2},x_{3},x_{4},x_{5}$ is
\begin{align}
    x_{2}-x_{3} &= I_{F}\cos{2\phi_{F}}  \\ 
    x_{4}+x_{5} &= I_{F}\sin{2\phi_{F}}  \\ 
    x_{2}+x_{3} &= I_{R}\cos{2\phi_{R}}  \\
    -x_{4}+x_{5} &= I_{R}\sin{2\phi_{R}}
\end{align}
then we can calculate the parameters as follows.

\noindent
Forward rotation component:
\begin{align}
    I_{F} &= \sqrt{(x_{2}-x_{3})^2+(x_{4}+x_{5})^2} \label{equ:I_F} \\
    \phi_{F} &= \frac{1}{2}\arctan{\frac{x_{4}+x_{5}}{x_{2}-x_{3}}} \label{equ:phi_F}
\end{align}
Reverse rotation component:
\begin{align}
    I_{R} &= \sqrt{(x_{2}+x_{3})^2+(-x_{4}+x_{5})^2} \label{equ:I_R}\\
    \phi_{R} &= \frac{1}{2}\arctan{\frac{-x_{4}+x_{5}}{x_{2}+x_{3}}} \label{equ:phi_R}
\end{align}
Unpolarized component:
\begin{equation}
    I_{U} = 2 x_{1} - I_{F} - I_{R}
\label{equ:I_U}
\end{equation}

The computational cost of the above algorithm is minimal because we can precompute the pseudo-inverse matrix $\mathbf{W}^{+}$ and easily parallelize per pixel. 

To solve Eq.~\ref{equ:I=WX}, the matrix $\mathbf{W}$ must have its own inverse matrix. Fortunately, if we choose the combinations of $(\theta_{c}, \theta_{l})$ carefully, $\mathbf{W}$ becomes invertible.
For example, $\mathbf{W}$ for a combination of $\{(\theta_{c}, \theta_{l})\} = \{(0, 0), (45,45), (0,45), (45,0), (90,0)\}$ is 
\begin{equation}
    \mathbf{W} =
    \begin{bmatrix}
    1 & 1 & 1 & 1 & 1 \\
    1 & 0 & 0 & 0 & -1 \\
    0 & 1 & 0 & 0 & 0 \\
    0 & 0 & 1 & 0 & 0 \\
    0 & 0 & 0 & 1 & 0 \\
    \end{bmatrix},
    \label{equ:W_case1}
\end{equation}
and this $\mathbf{W}$ is full-rank, so pseudo-inverse matrix $\mathbf{W}^{+}$ is calculable.

We also show the condition number of $\mathbf{W}$ to evaluate the robustness of our method. The condition number is defined by
\begin{equation}
    \kappa(\mathbf{W}) = \frac{\sigma_{\mathrm{max}}(\mathbf{W})}{\sigma_{\mathrm{min}}(\mathbf{W})},
\end{equation}
where $\sigma_{\mathrm{max}}(\mathbf{W})$ and $\sigma_{\mathrm{min}}(\mathbf{W})$ are maximal and minimal singular values of $\mathbf{W}$ respectively. If the condition number is close to one, the matrix $\mathbf{W}$ is considered well-conditioned and strong against noise. The condition number of the above scenario (Eq.~\ref{equ:W_case1}) is $\kappa=3.99$, which indicates that it is fairly well-conditioned. For more practical scenario where a polarization camera $\theta_{c}=\{0, 45, 90, 135\}$ captures two images with $\theta_{l}=\{0, 45\}$, the combination of polarizer angles is $\{(\theta_{c}, \theta_{l})\} = \{(0, 0), (45, 0), (90, 0), (135, 0),$ $(0, 45), (45, 45), (90, 45), (135, 45)\}$. In this case, the condition number becomes $\kappa=2.00$, indicating that it is well-conditioned. 

}


 





\end{document}